\documentclass[a4paper,fleqn]{cas-dc}

\usepackage[square,numbers]{natbib}
\usepackage{subfigure}
\usepackage{subcaption}
\usepackage{float}
\usepackage{pifont}

\def\tsc#1{\csdef{#1}{\textsc{\lowercase{#1}}\xspace}}
\tsc{WGM}
\tsc{QE}
\tsc{EP}
\tsc{PMS}
\tsc{BEC}
\tsc{DE}

\begin{document}
\let\WriteBookmarks\relax
\def\floatpagepagefraction{1}
\def\textpagefraction{.001}

\shorttitle{}

\shortauthors{Li et~al.}

\title[mode = title]{OBIFormer: A Fast Attentive Denoising Framework for Oracle Bone Inscriptions}                      



%
\author[1]{Jinhao Li}[style=chinese]



\ead{lomljhoax@stu.ecnu.edu.cn}


\affiliation[1]{organization={School of Computer Science and Technology, East China Normal University},
    city={Shanghai},
    postcode={200062}, 
    country={China}
    }

\author[2]{Zijian Chen}[style=chinese,orcid=0000-0002-8502-4110]
\ead{zijian.chen@sjtu.edu.cn}

\affiliation[2]{organization={Institute of Image Communication and Information Processing, Shanghai Jiao Tong University},
    city={Shanghai},
    postcode={200240}, 
    country={China}
    }

\author[3]{Tingzhu Chen}[style=chinese]
\cormark[1]
\ead{tingzhuchen@sjtu.edu.cn}

\credit{Data curation, Writing - Original draft preparation}

\affiliation[3]{organization={School of Humanities, Shanghai Jiao Tong University},
    city={Shanghai},
    postcode={200030}, 
    country={China}
    }

\author[4]{Zhiji Liu}[style=chinese]
\ead{251755227@qq.com}

\affiliation[4]{organization={Center for the Study and Application of Chinese Characters, East China Normal University},
    city={Shanghai},
    postcode={200241}, 
    country={China}
    }
    
\author[1]{Changbo Wang}[style=chinese,
                      auid=000,bioid=1]

\ead{cbwang@cs.ecnu.edu.cn}

\cortext[cor1]{Corresponding author.}


\begin{abstract}
Oracle bone inscriptions (OBIs) are the earliest known form of Chinese characters and serve as a valuable resource for research in anthropology and archaeology. However, most excavated fragments are severely degraded due to thousands of years of natural weathering, corrosion, and man-made destruction, making automatic OBI recognition extremely challenging. Previous methods either focus on pixel-level information or utilize vanilla transformers for glyph-based OBI denoising, which leads to tremendous computational overhead. Therefore, this paper proposes a fast attentive denoising framework for oracle bone inscriptions, i.e., OBIFormer. It leverages channel-wise self-attention, glyph extraction, and selective kernel feature fusion to reconstruct denoised images precisely while being computationally efficient. Our OBIFormer achieves state-of-the-art denoising performance for PSNR and SSIM metrics on synthetic and original OBI datasets. Furthermore, comprehensive experiments on a real oracle dataset demonstrate the great potential of our OBIFormer in assisting automatic OBI recognition. The code will be made available at \url{https://github.com/LJHolyGround/OBIFormer}.
\end{abstract}


\maketitle

\section{Introduction}

Oracle bone inscriptions (OBIs) represent China's earliest known mature and systematic writing system. Carved on materials such as oxen or turtle bones for divination and record-keeping during the late Shang and Zhou Dynasties, they provide profound insights into Chinese social culture for anthropologists and archaeologists. Therefore, the recognition of OBIs is an indispensable step in exploring the history of ancient China. However, despite the numerous fragments excavated, only a small percentage of these OBIs has been successfully recognized. Since the annotation task requires a high demand of time and labor with domain knowledge from OBI experts, there is an urgent need to develop an automatic OBI recognition algorithm. Unfortunately, many oracle bones have suffered considerable degradation over millennia due to natural weathering, corrosion, and man-made destruction, making automatic OBI recognition extremely challenging.

\begin{figure}[ht]
    \centering
    \includegraphics[width=\linewidth]{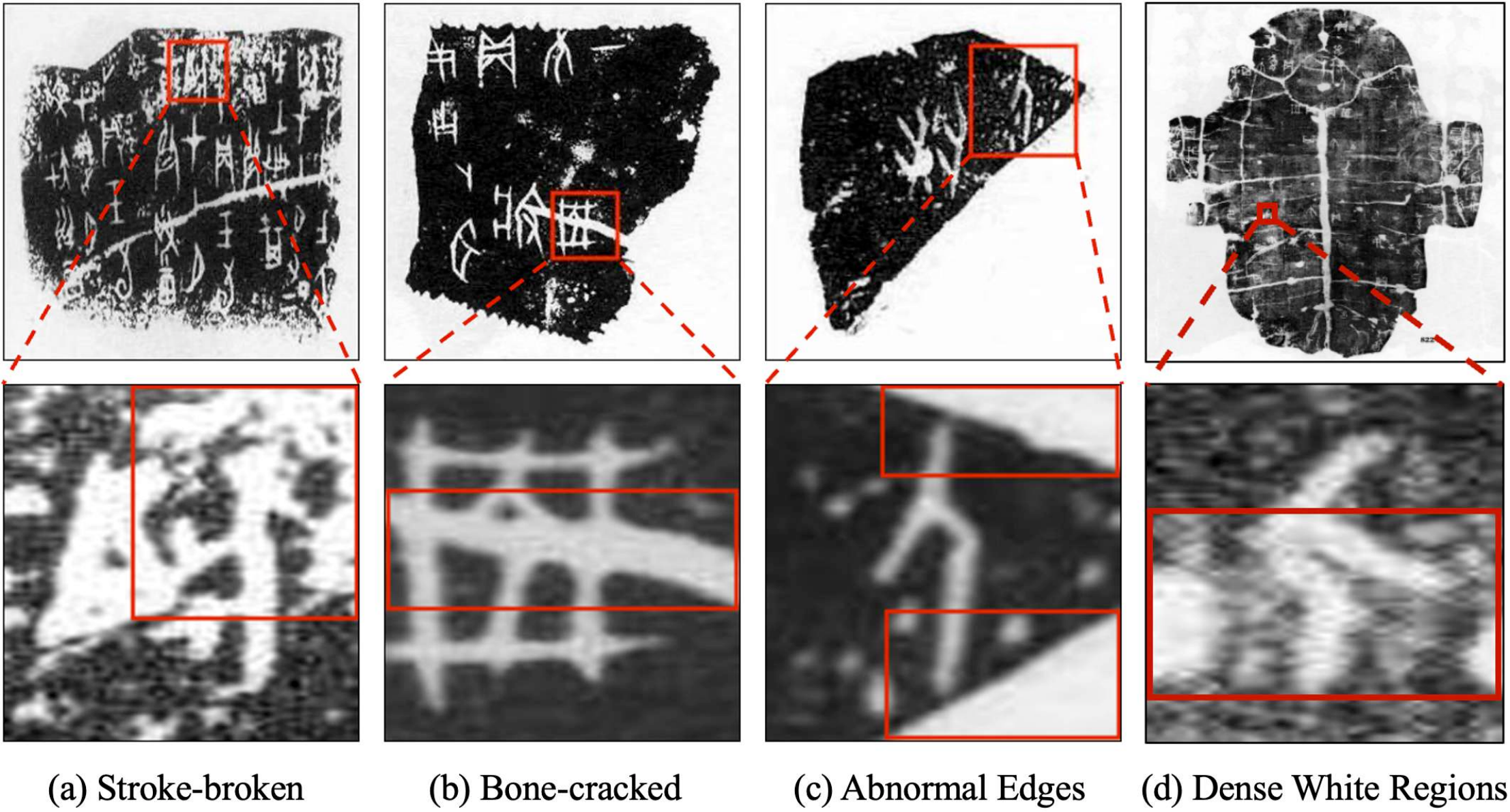}  
    \caption{Four types of noise in real rubbings. The red rectangle indicates the corresponding noise. \textbf{(a)} Stroke-broken, \textbf{(b)} Bone-cracked, \textbf{(c)} Abnormal edges, \textbf{(d)} Dense white regions.}
    \label{noises}
\end{figure}

As shown in Fig. \ref{noises}, noise in oracle bone inscriptions can be categorized into four types: stroke-broken, bone-cracked, abnormal edges, and dense white regions \cite{huang2019obc306}. Stroke-broken refers to the clusters of white regions near the strokes, complicating OBI recognition. Bone-cracked is caused by occlusion and typically runs through the center of the image, disrupting the glyph information. Abnormal edges also arise from occlusion and usually appear along the image boundaries. Besides, dense white regions represent fog-like noise, which obscures the structure and exacerbates ambiguity.

Traditional methods were first applied for OBI denoising. For example, Huang \emph{et al.} \cite{huang2016comparison} conducted a comprehensive comparative study of image denoising techniques relying on methods such as the anisotropic diffusion filter, Wiener filter, total variation, non-local means, and bilateral filtering. However, these methods do not perform well due to the complex degradation in OBIs. Therefore, some tailored denoising methods have been designed for the different noise types. Gu \emph{et al.} \cite{gu2010restoration} proposed an in-painting algorithm based on Poisson distribution and fractal geometry to remove erosion noise in OBIs. Khankasikam \cite{khankasikam2014restoration} proposed a binarization method based on adaptive multilayer information to restore uneven backgrounds in degraded historical document images. Robust Kronecker component analysis (RKCA) \cite{bahri2017robust, bahri2018robust} has recently been proposed for handling structured data with inherent tensorial properties. Subsequently, Zhang \emph{et al.} \cite{zhang2022kronecker} designed the Kronecker component with the low-rank dictionary (KCLD) and the Kronecker component with the robust low-rank dictionary (KCRD), which takes the Nuclear norm into RKCA to better capture the low-rank property of the two dictionaries in the basic sparse representation model. Considering the internal structural, spatial, and spectral information of the image block, they proposed the robust low-rank analysis with adaptive weighted tensor (AWTD) \cite{zhang2022robust}, a method that applies the adaptive weight tensor to the low-rank tensor model for image denoising.

With the prevalence of deep learning, numerous methods have been proposed for OBI denoising. For example, DnCNN \cite{zhang2017beyond} was first introduced to handle blind noise with unknown noise levels. Zamir \emph{et al.} \cite{zamir2022restormer} designed Restormer, an efficient encoder-decoder transformer for high-resolution image restoration. However, these generic denoising models mainly focus on pixel-level information while neglecting glyph information, which leads to poor performance. RCRN \cite{shi2022rcrn} first utilizes character skeleton information to achieve real-world character image restoration. Nevertheless, it extracts skeleton information from the input images, which results in unsatisfactory results. CharFormer \cite{shi2022charformer} attempts to leverage glyph information from the target images to address the problem, but it adopts a simple feature fusion strategy that limits its performance. Moreover, it leverages three vanilla transformer blocks in the residual self-attention block, which leads to tremendous computational overhead.

Additionally, automatic OBI recognition also suffers from data scarcity. In the early days, data augmentation methods primarily relied on techniques such as random rotation and flipping. Later, Han \emph{et al.} \cite{han2020self} introduced an Orc-Bert Augmentor pre-trained by self-supervised learning to recover masked input and generate pixel-format images as augmented data. Recently, Wang \emph{et al.} \cite{wang2022unsupervised} proposed a structure-texture separation network (STSN), which disentangles features into structure and texture components. It swaps texture information between any pairs of images so that transformed images are realistic and diverse. 

\begin{figure}[t]
    \centering
    \subfigure[PSNR on Oracle-50K dataset \cite{han2020self}]{
        \includegraphics[width=0.48\linewidth]{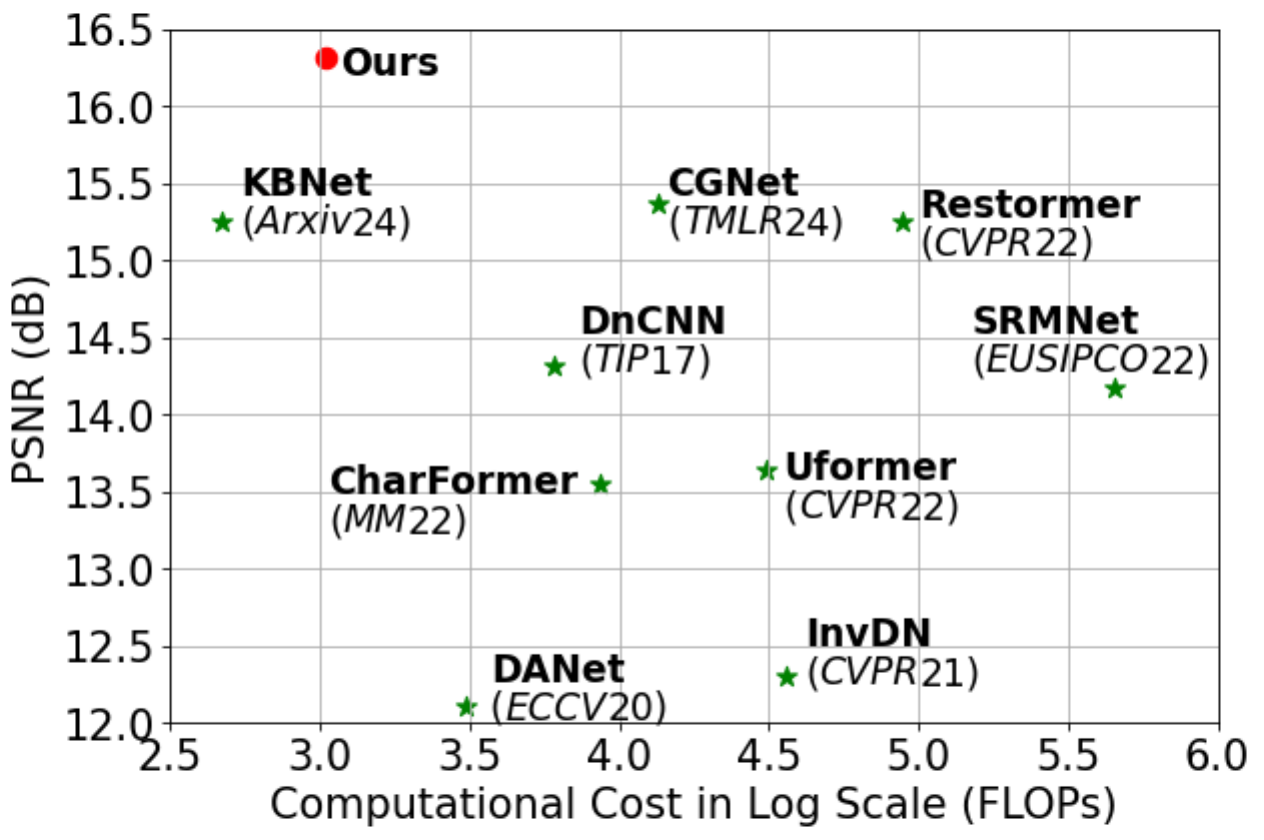}
        \hspace{-0.05\linewidth} 
    }
    \subfigure[SSIM on Oracle-50K dataset \cite{han2020self}]{
        \includegraphics[width=0.48\linewidth]{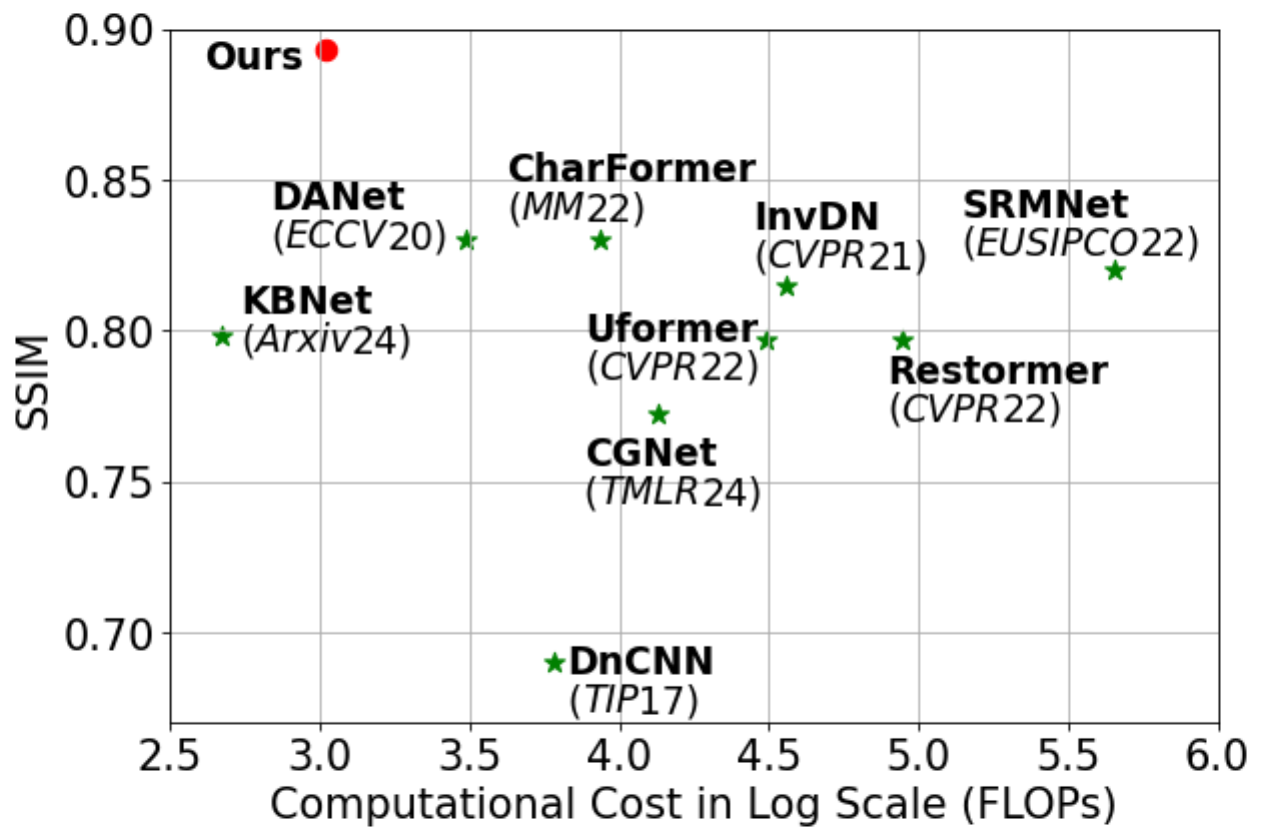}
    }
    \subfigure[PSNR on RCRN dataset \cite{shi2022rcrn}]{
        \includegraphics[width=0.48\linewidth]{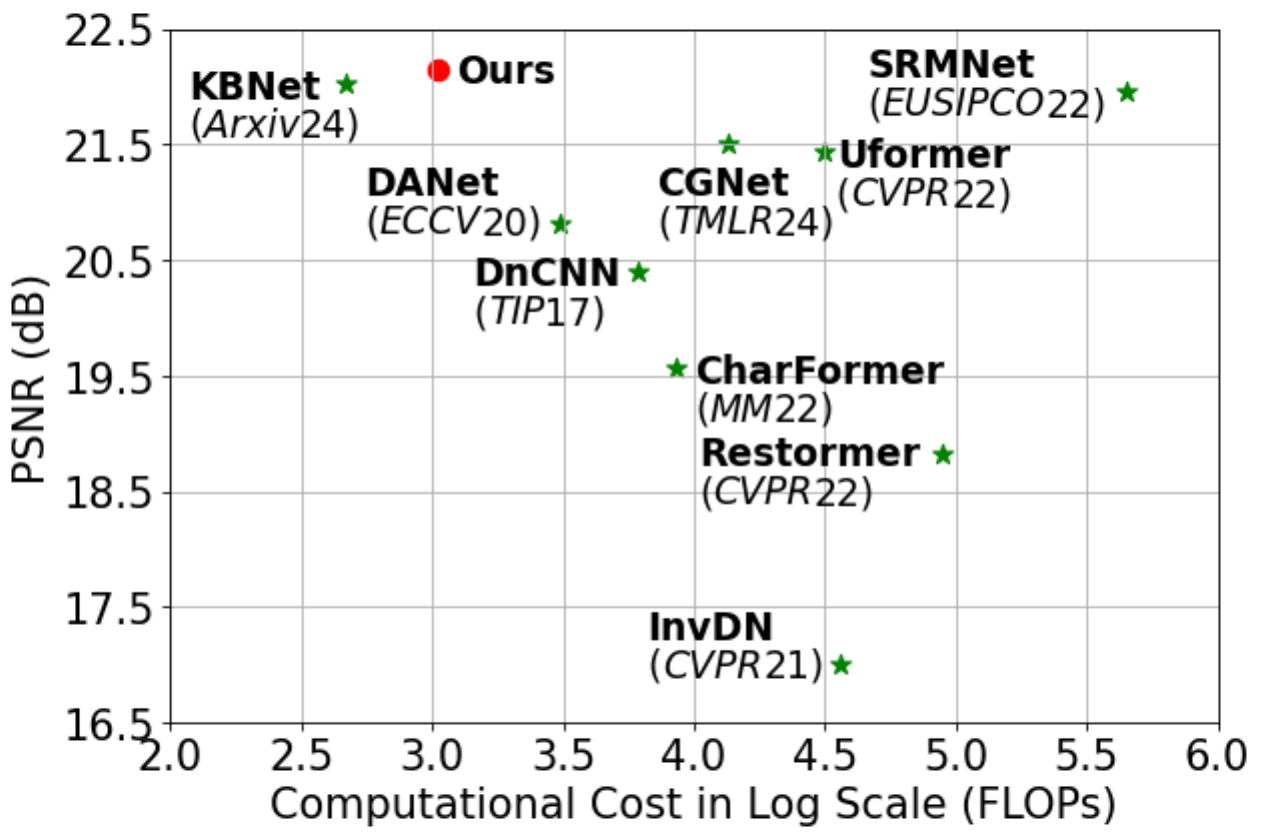}
        \hspace{-0.05\linewidth} 
    }
    \subfigure[SSIM on RCRN dataset \cite{shi2022rcrn}]{
        \includegraphics[width=0.48\linewidth]{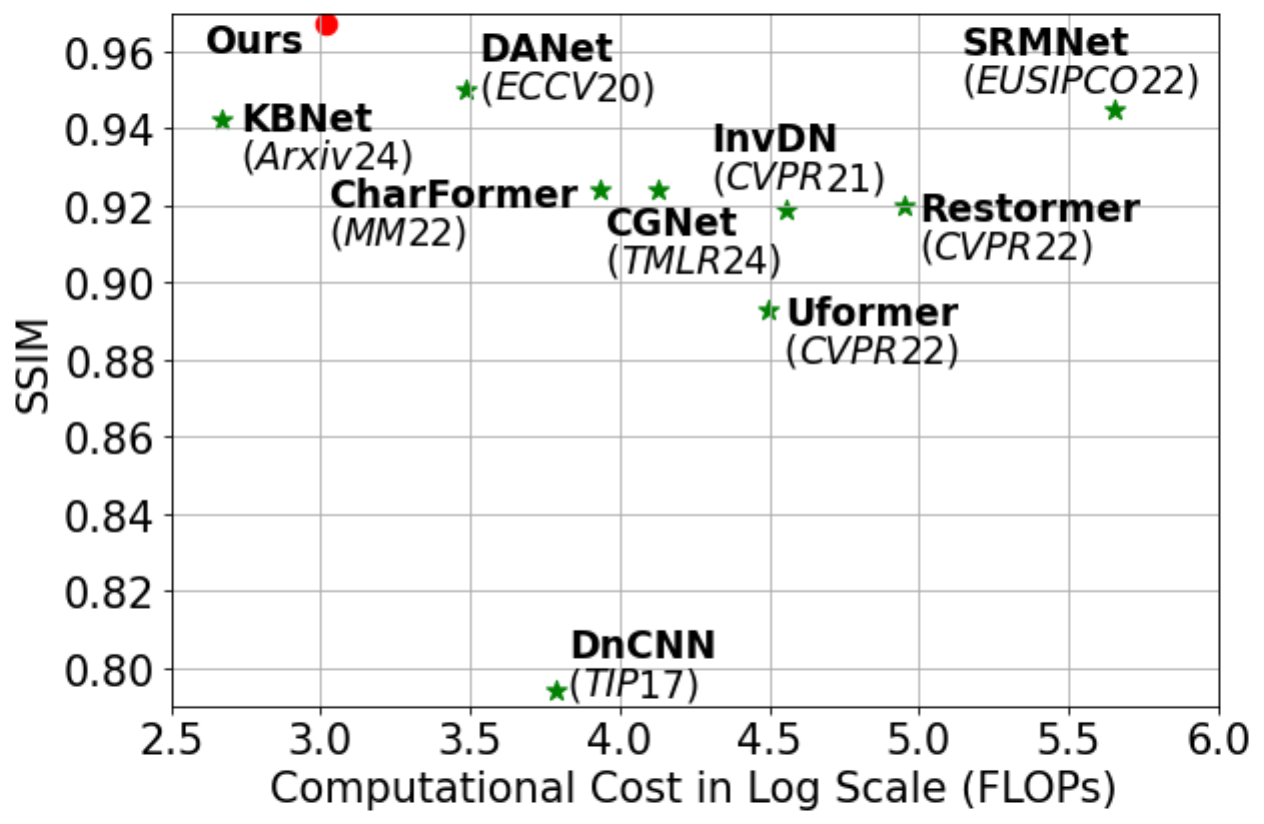}
    }
    \caption{Our model achieves state-of-the-art performance on the OBI denoising task while being computationally efficient.}
    \label{performance}
\end{figure}

To address these problems, we propose a fast attentive denoising framework for OBIs, i.e., OBIFormer, which utilizes channel-wise self-attention, glyph extraction, and selective kernel feature fusion. Specifically, our OBIFormer comprises an input projector, an output projector, an additional feature corrector, and several OBIFormer blocks (OFBs). The input projector extracts shallow features from the input images. The OBIFormer block is composed of channel-wise self-attention blocks (CSAB), glyph structural network blocks (GSNB), and a selective kernel feature fusion (SKFF) module. The SKFF module aggregates the reconstruction and glyph features captured by CSAB and GSNB. Finally, the output projector reconstructs the denoised image, and the additional feature corrector obtains the skeleton image. We conduct comprehensive experiments and demonstrate the effectiveness of our OBIFormer for OBI denoising tasks on Oracle-50K \cite{han2020self} and RCRN \cite{shi2022rcrn} datasets (See Fig. \ref{performance}). Furthermore, extensive experiments on a real oracle dataset (i.e., the OBC306 dataset \cite{huang2019obc306}) demonstrate its great potential in assisting automatic OBI recognition. Finally, we provide ablation studies to show the effectiveness of architectural designs and experimental choices. The main contributions of this work can be summarized as follows:
\begin{itemize}
    \item We propose a fast attentive denoising framework for OBIs, i.e., OBIFormer, based on channel-wise self-attention, glyph extraction, and selective kernel feature fusion. Additionally, we apply domain adaptation for the Oracle-50K dataset to synthesize noisy images for experiments.
    \item Comprehensive experiments on synthetic and original OBI datasets demonstrate the superiority of our OBIFormer for OBI denoising tasks. Furthermore, our OBIFormer is computationally efficient compared to other baseline methods.
    \item Extensive experiments on the OBC306 dataset show the strong generalization ability of the proposed OBIFormer. It demonstrates the great potential of OBIFormer in assisting automatic OBI recognition.
\end{itemize}

\section{Related Works}

\subsection{Oracle Bone Inscriptions Datasets}

\begin{figure*}[ht]
    \centering
    \subfigure[Oracle-50K \cite{han2020self}]{
        \includegraphics[width=0.31\textwidth]{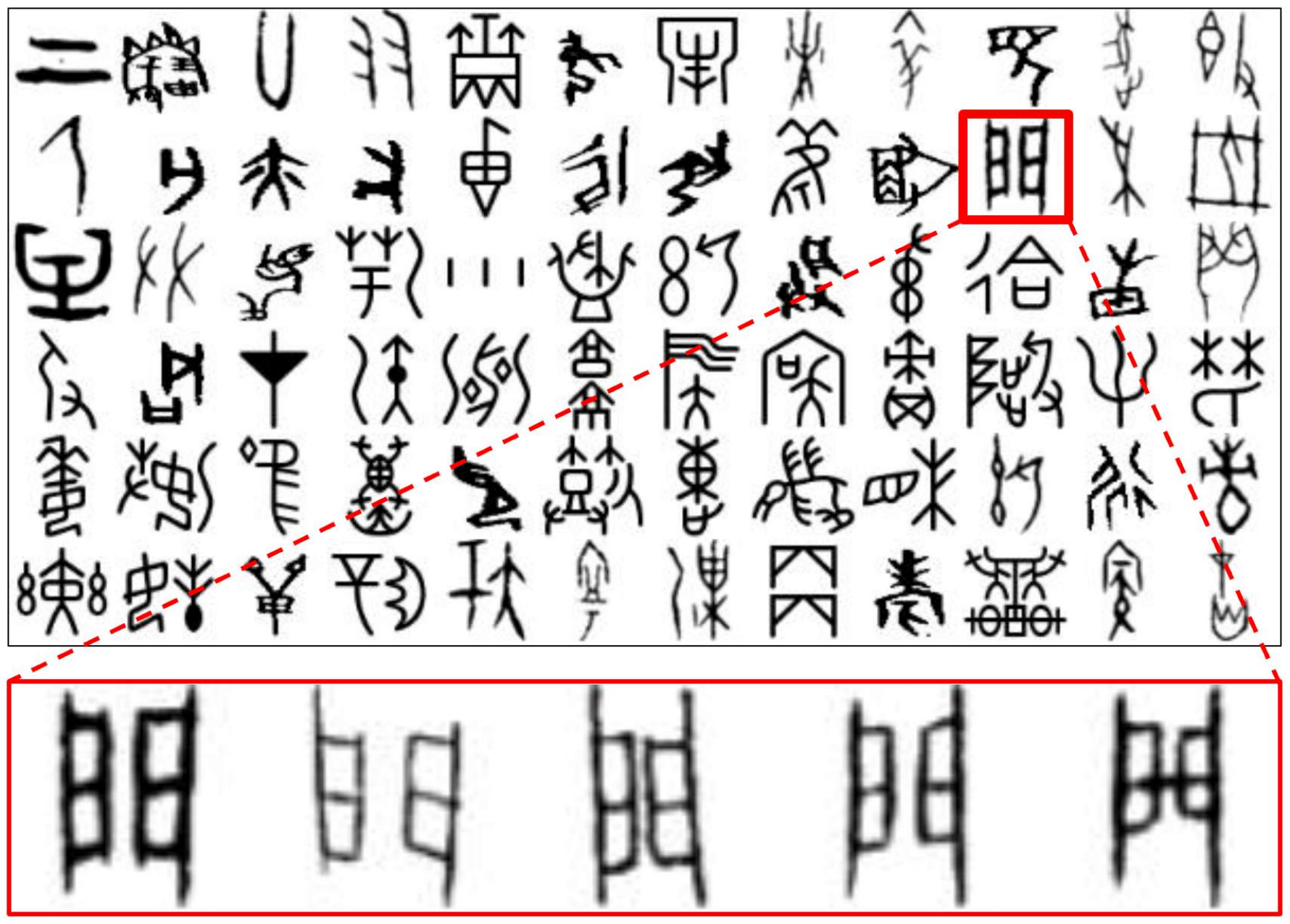}
    }
    \subfigure[HWOBC \cite{li2020hwobc}]{
        \includegraphics[width=0.31\textwidth]{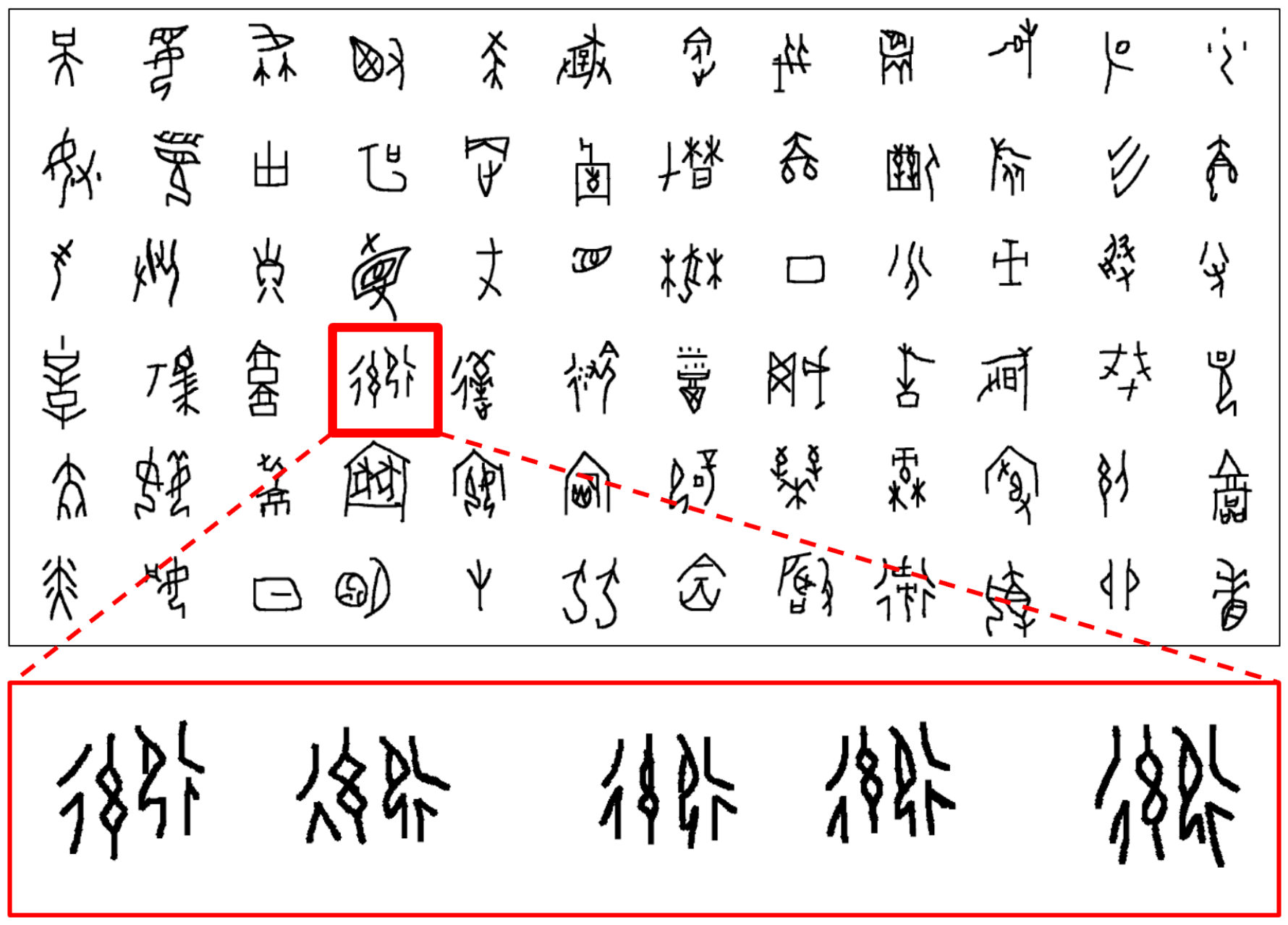}
    }
    \subfigure[EVOBI \cite{wang2022study}]{
        \includegraphics[width=0.31\textwidth]{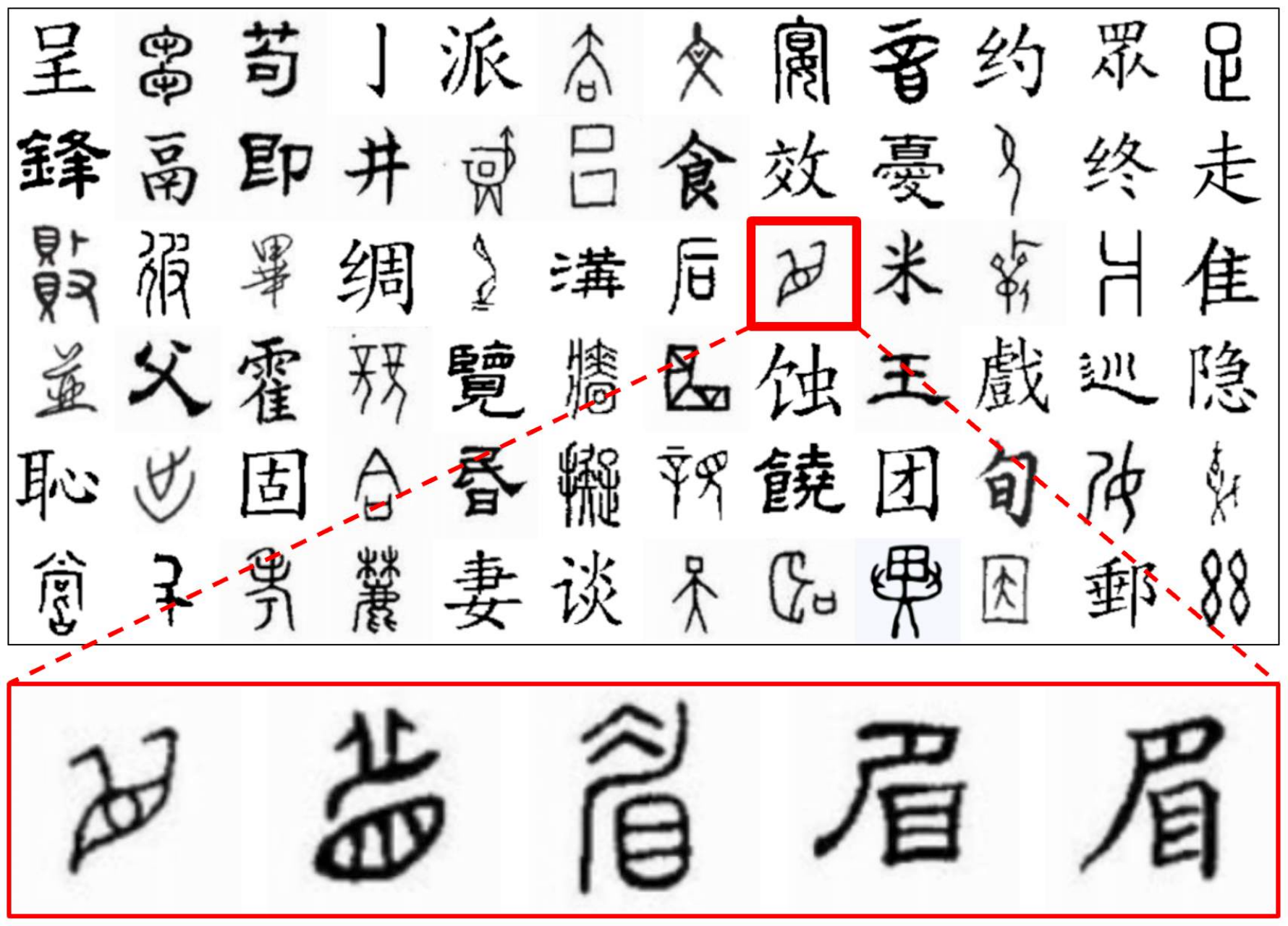}
    }
    \subfigure[OBC306 \cite{huang2019obc306}]{
        \includegraphics[width=0.31\textwidth]{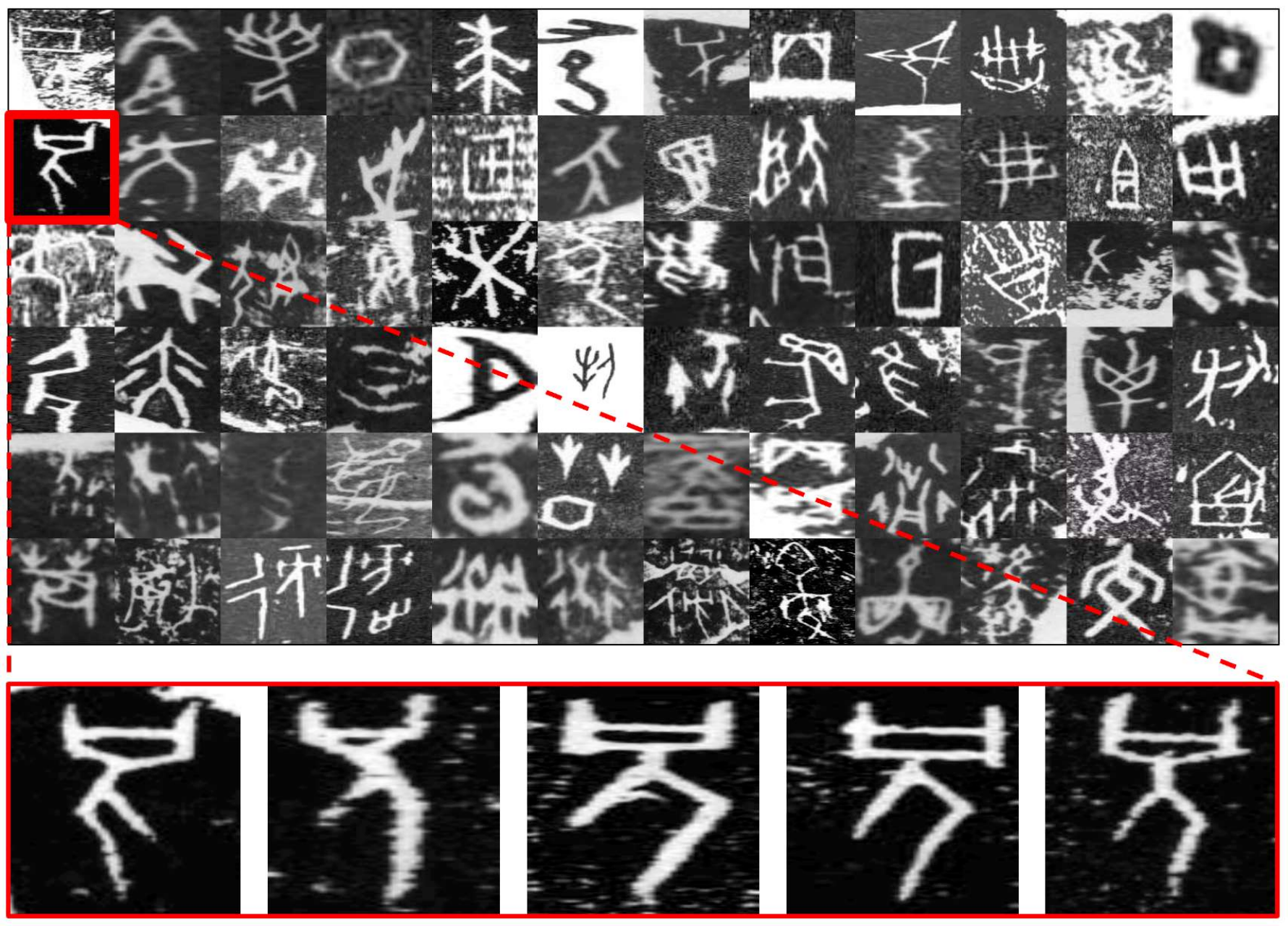}
    }
    \subfigure[OBI125 \cite{yue2022dynamic}]{
        \includegraphics[width=0.31\textwidth]{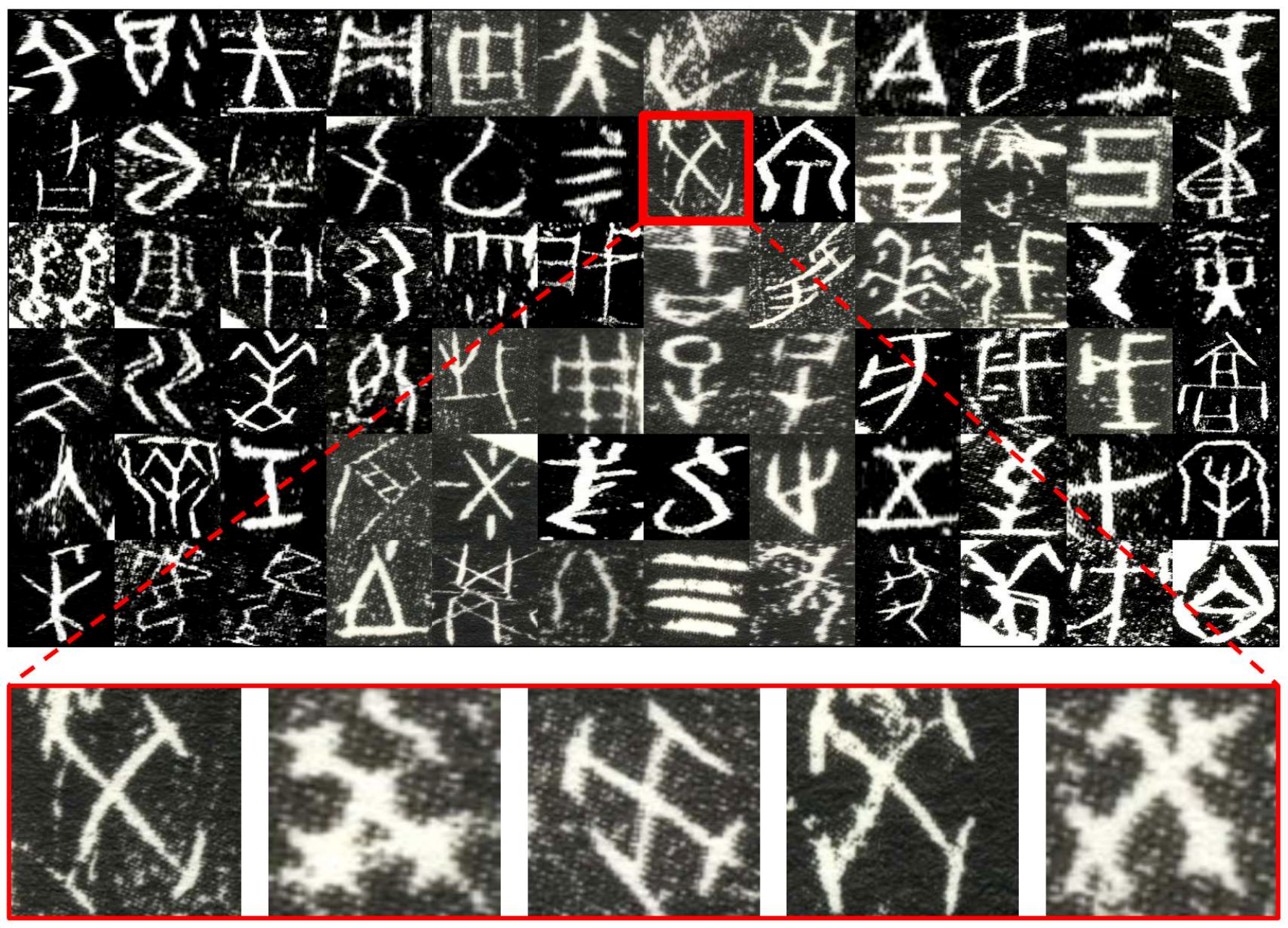}
    }
    \subfigure[EVOBC \cite{guan2024open}]{
        \includegraphics[width=0.31\textwidth]{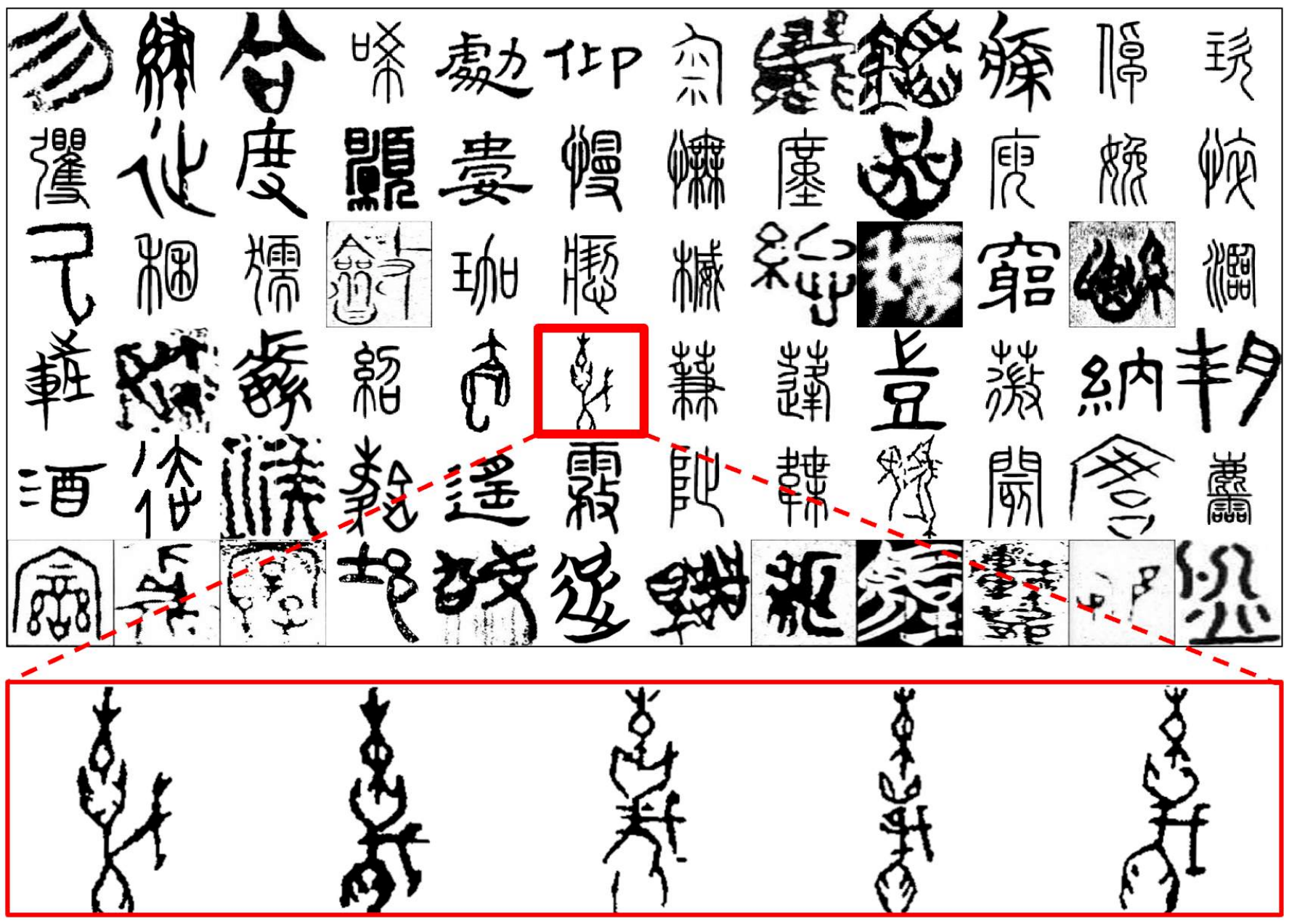}
    }
    \caption{Examples of oracle character images in different OBI datasets: \textbf{(a)} Oracle-50K \cite{han2020self}, \textbf{(b)} HWOBC \cite{li2020hwobc}, \textbf{(c)} EVOBI \cite{wang2022study}, \textbf{(d)} OBC306 \cite{huang2019obc306}, \textbf{(e)} OBI125 \cite{yue2022dynamic}, and \textbf{(f)} EVOBC dataset \cite{guan2024open}. The zoomed-in images are different structural variations of the same character.}
    \label{samples}
\end{figure*}

In recent years, various OBI datasets have been established for OBI-related computer vision tasks, such as oracle bone recognition, rejoining, classification, retrieval, and deciphering \cite{chen2025obibenchlmmsaidstudy}. From the perspective of the content sources, OBI datasets can be categorized into handprints and rubbings, as shown in Fig. \ref{samples}. Furthermore, we summarize their statistical information in Table \ref{statistic}.

\begin{table*}[ht]
    \caption{Summary of the existing oracle bone inscription datasets.}
    \label{statistic}
    \begin{tabular*}{\tblwidth}{@{}CCCCCCCC@{}}
	\toprule
	\toprule
	Type & Dataset & Year & \#Classes & \#Total & Avg. Res. & Application & Availability \\ 
        \midrule
	\multirow{7}{*}{Handprints} & Oracle-20K \cite{guo2015building} & 2016 & 261 & 20,039 & 50$\times$50 & Recognition & {\color{green}\ding{51}} \\
	 & Yang \emph{et al.} \cite{yang2018accurate}  & 2018 & 39 & 21,373 & - & Recognition & {\color{red}\ding{55}} \\
	 & Liu \emph{et al.} \cite{du2021deep} & 2018 & 5,491 & 44,868 & - & Recognition & {\color{red}\ding{55}} \\
        & Oracle-50K \cite{han2020self} & 2020 & 2,668 & 59,081 & 50$\times$50 & Recognition & {\color{green}\ding{51}} \\
        & HWOBC \cite{li2020hwobc} & 2020 & 3,881 & 83,245 & 400$\times$400 & Recognition & {\color{green}\ding{51}} \\
        & EVOBI \cite{wang2022study} & 2022 & 972 & 4,860 & 105$\times$105 & Recognition & {\color{green}\ding{51}} \\
        \midrule
	\multirow{12}{*}{Rubbings} & OBC306 \cite{huang2019obc306} & 2019 & 306 & 309,551 & <382$\times$478 & Recognition & {\color{green}\ding{51}} \\
        & Wang \emph{et al.} \cite{1019185421.nh} & 2019 & - & 7,824 & - & Denoising \& Recognition & {\color{red}\ding{55}} \\
        & Liu \emph{et al.} \cite{liu2020oracle} & 2020 & 682 & - & - & Recognition & {\color{red}\ding{55}} \\
        & OracleBone8000 \cite{zhang2021ai} & 2020 & - & 129,770 & - & Rejoining \& Recognition & {\color{red}\ding{55}} \\
        & Yoshiyuki \emph{et al.} \cite{fujikawa2023recognition} & 2022 & 27 & 649 & - & Detection \& Recognition & {\color{red}\ding{55}} \\
        & OBI125 \cite{yue2022dynamic} & 2022 & 125 & 4,257 & <278$\times$473 & Recognition & {\color{green}\ding{51}} \\
        & OB-Rejoin \cite{zhang2022data} & 2022 & - & 998 & <1408$\times$1049 & Rejoining & {\color{red}\ding{55}} \\
        & Oracle-MINST \cite{wang2024dataset} & 2023 & 10 & 30,222 & 28$\times$28 & Recognition & {\color{green}\ding{51}} \\
        & EVOBC \cite{guan2024open} & 2024 & 13,714 & 229,170 & <465$\times$857 & Generation \& Recognition & {\color{green}\ding{51}} \\
        & HUST-OBC \cite{wang2024open} & 2024 & 10,999 & 140,053 & <400$\times$520 & Recognition & {\color{green}\ding{51}} \\
        & O2BR \cite{chen2025obibenchlmmsaidstudy} & 2025 & - & 800 & $<$2664$\times$2167 & Recognition & {\color{green}\ding{51}} \\
        & OBI-rejoin \cite{chen2025obibenchlmmsaidstudy} & 2025 & - & 200 & $<$2913$\times$1268 & Rejoining & {\color{green}\ding{51}} \\
        \midrule
        \multirow{2}{*}{Hybrid} & Oracle-241 \cite{wang2022unsupervised} & 2022 & 241 & 78,565 & <588$\times$700 & Generation \& Recognition & {\color{green}\ding{51}} \\
        & Oracle-P15K \cite{li2025mitigatinglongtaildistributionoracle} & 2025 & 239 & 14,542 & 128$\times$128 & Generation \& Denoising & {\color{green}\ding{51}} \\
	\bottomrule
	\bottomrule
\end{tabular*}
\end{table*}

Handprinted OBI datasets contain distortion-free rubbing images rewritten by OBI experts. For example, the Oracle-20K dataset \cite{guo2015building} was collected and labeled by Guo \emph{et al.} from the Chinese etymology website \footnote{http://www.chineseetymology.org/}. It encompasses 20,039 oracle character instances across 261 categories. Later, Han \emph{et al.} \cite{han2020self} further expanded the number of instances and proposed the Oracle-50K dataset, which consists of 59,081 instances belonging to 2,668 categories. Similarly, Li \emph{et al.} built the HWOBC dataset \cite{li2020hwobc} for training automatic OBI recognition models. It became the largest handprinted OBI dataset with 83,245 character-level samples grouped into 3,881 categories. Moreover, the EVOBI dataset \cite{wang2022study} was proposed for OBI interpretation, which was crawled from the Chinese master website \footnote{http://www.guoxuedashi.net/}. Nevertheless, these datasets only provide handprinted images, contributing minimally to automatic OBI recognition. To synthesize more realistic OBI data, Wang \emph{et al.} \cite{wang2022unsupervised} constructed the Oracle-241 dataset to transfer knowledge across handprinted characters and scanned data. It comprises 78,565 handprinted and scanned images of 241 categories. Most recently, Li \emph{et al.} \cite{li2025mitigatinglongtaildistributionoracle} proposed Oracle-P15K, a structure-aligned OBI dataset consisting of 14,542 images infused with domain knowledge from OBI experts, to achieve realistic and controllable OBI generation. Comprehensive experiments demonstrate that the augmented images can mitigate the long-tail distribution problem in existing OBI datasets.

Rubbing OBI datasets consist of scanned images from oracle bone publications, most of which suffer from severe and distinctive noise caused by thousands of years of natural weathering, corrosion, and man-made destruction. One of the representative datasets, OBC306 \cite{huang2019obc306}, encompasses 309,551 samples classified into 306 classes from eight authoritative oracle bone publications. The OracleBone8000 dataset \cite{zhang2021ai} contains 129,770 images with character-level annotations. However, it is highly imbalanced and sparse, limiting itself to serving as a comprehensive benchmark for automatic OBI recognition tasks. Yue \emph{et al.} \cite{yue2022dynamic} built the OBI125 dataset, which is composed of 4,257 images scanned from the collection of the Shanghai Museum (Volume I) \cite{pu2009oracle}. For the oracle bone rejoining task, Zhang \emph{et al.} \cite{zhang2022data} proposed the OB-Rejoin dataset, which covers different writing styles and fonts, featuring 249 pairs of known rejoining manually found by OBI experts over the past few decades. Moreover, the Oracle-MINST dataset \cite{wang2024dataset} was released to benchmark the oracle bone character classification task, following the same data format as the well-known MNIST dataset \cite{lecun1998gradient}. Similar to EVOBI, the EVOBC dataset \cite{guan2024open} was constructed to trace the evolution of Chinese characters from oracle bone inscriptions to their contemporary forms, including written representations of the same character across different historical periods. Wang \emph{et al.} \cite{wang2024open} introduced the HUST-OBC dataset, which comprises 141,053 images sourced from five different origins. Among them, 77,064 images spanning 1,781 categories have been deciphered, while 62,989 images across 9,411 categories remain undeciphered. Most recently,  Chen \emph{et al.} \cite{chen2025obibenchlmmsaidstudy} proposed O2BR and OBI-rejoin datasets to evaluate the recent large multi-modal models (LMMs) in OBI recognition and rejoining tasks. Each image in these datasets is equipped with several questions alongside correct answers.
 
\subsection{Oracle Bone Inscription Recognition}

The traditional OBI recognition adopts a three-stage pipeline paradigm: data pre-processing, feature extraction, and recognition. In the early days, Guo \emph{et al.} \cite{guo2015building} proposed a hierarchical representation that integrates a Gabor-related low-level representation and a sparse-encoder-related mid-level representation with CNN-based models, which achieved better performance over both approaches. Similarly, Yang \emph{et al.} \cite{yang2018accurate} also applied a feature extraction technique in CNN, but CNN completes both feature extraction and OBI recognition tasks. However, the simple feature representation limits the performance of traditional OBI recognition methods. Later, Du \emph{et al.} \cite{du2021deep} designed a two-branch deep learning framework consisting of two pretext tasks for rotation and deformation. Experimental results demonstrated that their method could effectively learn features of OBIs and provide good feature representation for the downstream tasks.

Though achieving good accuracy, most traditional methods rely heavily on manually designed complex features at multiple levels. Therefore, they are not suitable for dealing with large-scale datasets. In contrast, deep learning methods are known for their representation capacity when processing large-scale datasets. Huang \emph{et al.} \cite{huang2019obc306} first introduced the standard deep CNN-based evaluation (i.e., AlexNet \cite{krizhevsky2012imagenet}, Inception-v4 \cite{szegedy2017inception}, VGG16 \cite{simonyan2014very}, ResNet-50, and ResNet-101 \cite{he2016deep}) for the OBC306 dataset, which served as a benchmark. However, the long-tail distribution problem in OBI datasets hampers the performance of deep learning methods. Hence, Zhang \emph{et al.} \cite{zhang2019oracle} uses a CNN to map the character images to a Euclidean space where the nearest neighbor rule classifies them. Most recently, Wang \emph{et al.} \cite{wang2022unsupervised} proposed a structure-texture separation network (STSN), which separates texture from the structure to avoid the negative influence caused by degradation. Nevertheless, the recognition accuracy of scanned images is still unsatisfactory. Afterward, they trained an unsupervised discriminative consistency network (UDCN) \cite{wang2024oracle} with an unsupervised transition loss and leveraged pseudo-labeling to make the predictions of scanned samples consistent under different noise.

\subsection{Image Denoising}

Image denoising plays a vital role in different image processing applications, such as medical imaging \cite{prada2024statistical, chen2021lesion}, remote sensing \cite{liu2023indeandcoe}, and oracle bone research \cite{shi2022rcrn, shi2022charformer}. Since the rise of deep learning, researchers have proposed various models for image denoising. For example, Zhang \emph{et al.} \cite{zhang2017beyond} leveraged residual learning and batch normalization to introduce DnCNN, a model capable of handling blind noise with unknown noise levels. Later, Fan \emph{et al.} \cite{fan2022selective} presented SRMNet, a blind real-image denoising network utilizing a hierarchical architecture improved from U-Net \cite{ronneberger2015u}, which advanced the performance on two synthetic and two real-world noisy datasets. However, it requires tremendous computational overhead due to the complex hierarchical architecture. Therefore, Zhang \emph{et al.} \cite{zhang2023kbnet} proposed KBNet, which consists of a kernel basis attention module and a multi-axis feature fusion block. It performs state-of-the-art on over ten image denoising benchmarks while maintaining low computational overhead. Besides, Liu \emph{et al.} \cite{liu2021invertible} designed a lightweight, information-lossless, and memory-saving invertible neural network (INN) based model, namely InvDN, which replaces the noisy latent representation with another one sampled from a prior distribution during reversion and achieves promising results.

Generative adversarial network (GAN) based models have recently been applied for image denoising. Yue \emph{et al.} \cite{yue2020dual} introduced DANet, a unified framework synthesizing noisy-clean image pairs simultaneously with two new metrics. To preserve the structural consistency of characters, Shi \emph{et al.} \cite{shi2022rcrn} proposed RCRN, which comprises a skeleton extractor (SENet) and a character image restorer (CiRNet). It is capable of handling complex degradation and specific noise types in OBIs.

More recently, researchers have introduced visual transformer (ViT) \cite{dosovitskiy2021an} to image denoising tasks. Shi \emph{et al.} \cite{shi2022charformer} designed a glyph-based attentive framework, i.e., CharFormer, which maintains critical features of a character during the denoising process. Later, Wang \emph{et al.} \cite{wang2022uformer} proposed Uformer based on a locally-refined window (LeWin) transformer and a learnable multi-scale restoration modulator. It excels at capturing local and global dependencies, making it highly effective for image restoration. However, though these transformer-based models obtain impressive performance, their computational complexity grows quadratically with the spatial resolution. Therefore, Zamir \cite{zamir2022restormer} applied multi-Dconv head transposed attention (MDTA) and gated-Dconv feed-forward network (GDFN) in Restormer to balance the performance and computational overhead. Most recently, Ghasemabadi \emph{et al.} \cite{ghasemabadi2024cascadedgaze} presented CGNet to capture global information without self-attention, thus reducing the computational complexity significantly while maintaining outstanding performance.

\section{Method}

In this section, we elaborate on the overall pipeline and the hierarchical structure of our OBIFormer. Specifically, we provide the details of the OBIFormer block (OFB), which consists of channel-wise self-attention blocks (CSABs), glyph structural network blocks (GSNBs), and a selective kernel feature fusion (SKFF) module. The SKFF injects glyph features extracted by GSNBs into the denoising backbone, thereby guiding the model in removing the complex noise while preserving the inherent glyphs.

\subsection{Overall Pipeline}

\begin{figure*}[ht]
    \centering
    \includegraphics[width=\linewidth]{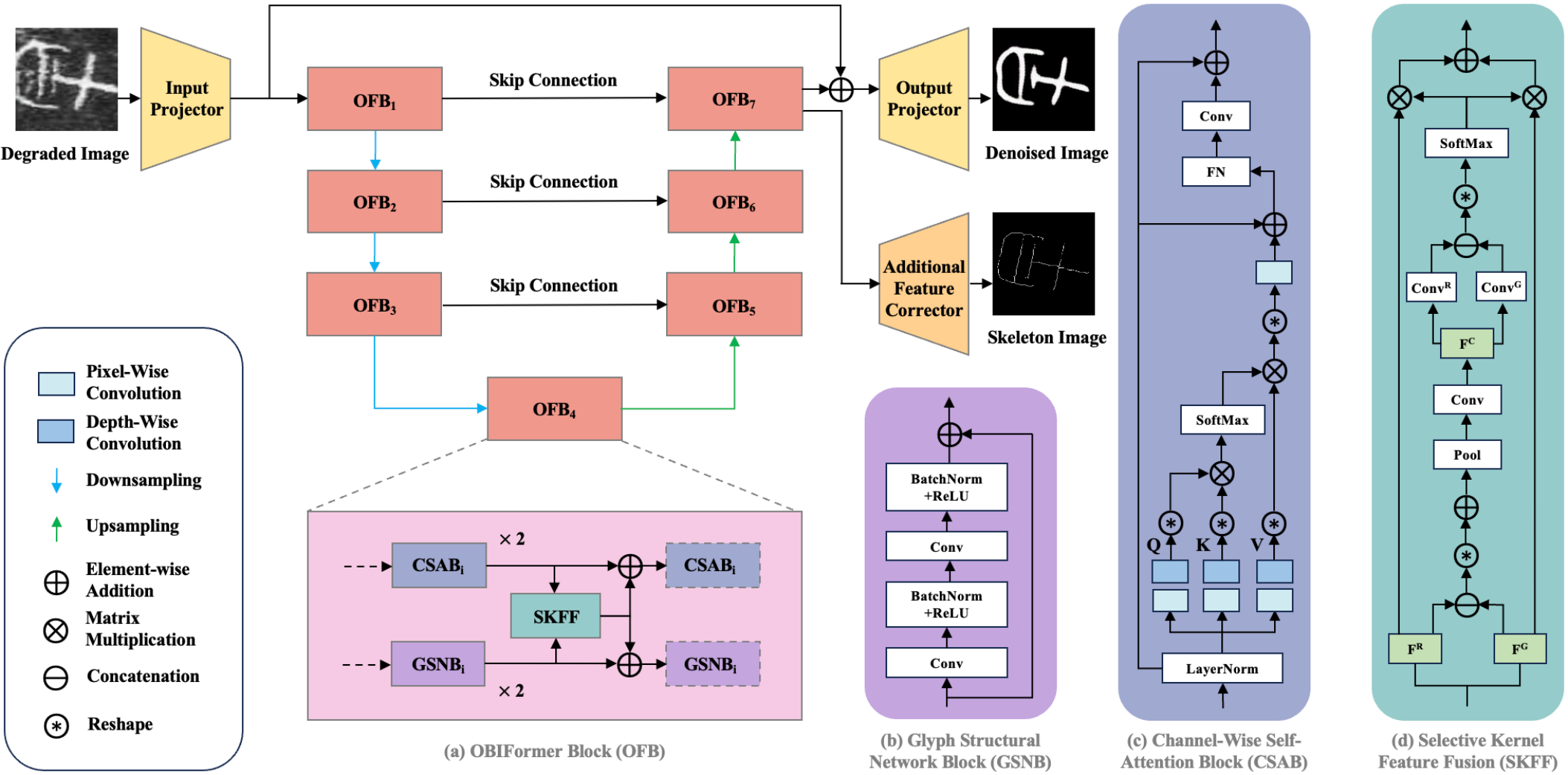}
    \caption{The overall architecture of our OBIFormer. \textbf{(a)} OBIFormer block (OFB) that injects glyph information into the denoising backbone, \textbf{(b)} Glyph structural network block (GSNB) that extracts glyph features, \textbf{(c)} Channel-wise self-attention block (CSAB) that generates channel-wise self-attention effectively and efficiently, \textbf{(d)} Selective kernel feature fusion (SKFF) module that aggregates reconstruction features and glyph features.}
    \label{arch}
\end{figure*}

As shown in Fig. \ref{arch}, the overall structure of our OBIFormer is a U-shaped encoder-decoder network with skip connections between the OFBs. Specifically, given a degraded image $\textbf{I}\in \mathbb{R}^{3\times H\times W}$, the input projector applies a $3\times 3$ convolution layer with LeakyReLU to extract the shallow features $\textbf{F}_0\in \mathbb{R}^{C\times H\times W}$ from the input image. Then, the shallow features $\textbf{F}_0$ are passed through $N+1$ encoder stages. Each stage contains an OFB and a downsampling layer. The OFB is designed for precise OBI image denoising based on the glyph feature extraction and aggregation. Given intermediate shallow features $\textbf{F}_{i}$, the output of ${\rm OFB}_{i}$ is:
\begin{equation}
    \textbf{F}_{i+1} = {\rm OFB}_{i+1}(\textbf{F}_{i}), 0\leq i \leq N.
\end{equation}
In the downsampling layer, we downsample the maps and double the channels using $4\times 4$ convolution with stride 2.

Next, the latent features $\textbf{F}_{N+1}$ are passed through $N$ decoder stages. Each stage contains an OFB and an upsampling layer. Moreover, we apply skip connections between the OFBs in the encoder and decoder to transmit extraction results on different scales and restore spatial features. Therefore, the deep features $\textbf{F}_{i+1}$ can be formulated as:
\begin{equation}
    \textbf{F}_{i+1} = {\rm OFB}_{i}(\textbf{F}_{i}+\textbf{F}_{2N-i+1}), N< i \leq 2N
\end{equation}
where $n$ is the stage number of the decoder, ${\rm OFB}_{2N-i+1}$ represents the $(2N-i+1)$-th OFB. For feature upsampling, we apply a transposed convolution operation, which upsamples the maps and halves channels using $2\times 2$ convolution with stride 2.

The last OFB produces $\textbf{F}_{2N+1}$, which consists of reconstruction features $\textbf{F}_{2N+1}^R$ and glyph features $\textbf{F}_{2N+1}^G$. Finally, reconstruction features $\textbf{F}_{2N+1}^R$ are aggregated with the shallow features $\textbf{F}_0^R$. The output projector reconstructs the restored image $\textbf{I}^{\prime}$ with a $3\times 3$ convolution layer:
\begin{equation}
    \textbf{I}^{\prime} = {\rm Conv}(\textbf{F}_0^R + \textbf{F}_{2N+1}^R).
\end{equation}
Similarly, the additional feature corrector adjusts the glyph features $\textbf{F}_{2N+1}^G$ to obtain the restored skeleton $\textbf{S}^{\prime}$ with a $3\times 3$ convolution layer:
\begin{equation}
    \textbf{S}^{\prime} = {\rm Conv}(\textbf{F}_{2N+1}^G).
\end{equation}

\subsection{OBIFormer Block}

As shown in Fig. \ref{arch}(a), the OBIFormer block (OFB) consists of two residual channel-wise self-attention blocks (CSABs) as the denoising backbone, two glyph structural network blocks (GSNBs) for glyph extraction, and a selective kernel feature fusion (SKFF) module for feature aggregation. The primary purpose of the OFB is to alleviate the computational overhead and aggregate glyph features. We elaborate on the designs of these modules as follows:

\noindent \textbf{Residual Channel-Wise Self-Attention Block.} The residual channel-wise self-attention block (CSAB) is a residual block with a modified transformer layer, as shown in Fig. \ref{arch}(c). Recently, transformer-based models have swept over various tasks with impressive results. However, the complexity of self-attention grows quadratically with the spatial resolution. Therefore, we apply channel-wise self-attention instead of spatial-wise self-attention, which remains effective while being computationally efficient. Specifically, each transformer layer performs channel-wise self-attention (CSA) and feed-forward (FN). For a normalized tensor $\textbf{F}_N$, query (\textbf{Q}), key (\textbf{K}), and value (\textbf{V}) are generated by applying $1\times 1$ convolutions to aggregate pixel-wise cross-channel context followed by $3\times 3$ depth-wise convolutions to encode channel-wise spatial context:
\begin{align}
    \textbf{Q} &= \textbf{W}_d^Q \textbf{W}_p^Q\textbf{F}_N, \\
    \textbf{K} &= \textbf{W}_d^K \textbf{W}_p^K\textbf{F}_N, \\
    \textbf{V} &= \textbf{W}_d^V \textbf{W}_p^V\textbf{F}_N,
\end{align}
where $\textbf{W}_d^{(\cdot)}$ and $\textbf{W}_p^{(\cdot)}$ denote the projection matrices obtained by $1\times 1$ point-wise convolution and $3\times 3$ depth-wise convolution with no bias, respectively. Then, we reshape the query and key projection so that their dot product generates a transposed-attention map of size $\mathbb{R}^{C\times C}$ instead of the regular attention map of size $\mathbb{R}^{HW\times HW}$:
\begin{align}
    {\rm Attention}(\textbf{Q}, \textbf{K}, \textbf{V})={\rm SoftMax}(\frac{\textbf{KQ}}{\alpha})\textbf{V},
\end{align}
where $\textbf{Q}\in \mathbb{R}^{HW\times C}, \textbf{K}\in \mathbb{R}^{C\times HW}$, and $\textbf{V}\in \mathbb{R}^{HW\times C}$ matrices are obtained after reshaping. $\alpha$ is a learnable scaling parameter. Therefore, the output features of the transformer layer in CSAB can be formulated as follows:
\begin{align}
    \textbf{F}_l^{\prime} &= \textbf{W}_p{\rm CSA}({\rm LN(\textbf{F}_{l-1}})) + \textbf{F}_{l-1}, \\
    \textbf{F}_l &= {\rm FN}({\rm LN}(\textbf{F}_l^{\prime})),
\end{align}
where LN represents layer normalization. $\textbf{F}_l^{\prime}$ refers to the intermediate output features of the $l$-th transformer layer. $\textbf{F}_{l-1}$ and $\textbf{F}_l$ denote the final output features of the ($l-1$)-th and $l$-th transformer layer, respectively. 

\noindent \textbf{Glyph Structural Network Block.} As shown in Fig. \ref{arch}(b), the glyph structural network block (GSNB) consists of two $3\times 3$ convolution layers with batch normalization and ReLU activation function in a residual block. Due to the strong capacity to capture local dependencies of CNNs, GSNB can effectively extract glyph features that will be fused with reconstruction features. Similarly, GSNB incorporates downsampling and upsampling layers to maintain the same scale as the corresponding RSAB. Given input glyph features $\textbf{F}_i^G\in\mathbb{R}^{C\times H\times W}$, the GSNB in $n$-th OFB will output $\textbf{F}_{i+1}^G$ which has the same size as $\textbf{F}_{i+1}^R$.

\noindent \textbf{Selective Kernel Feature Fusion.} To better aggregate the glyph features, we perform selective kernel feature fusion (SKFF) \cite{li2019selective} instead of simple addition or concatenation. As shown in Fig. \ref{arch}(d), the SKFF module dynamically adjusts the receptive field via a triplet of operators: $Split$, $Fuse$, and $Select$. The $Split$ operation generates reconstruction features $\textbf{F}^R$ and glyph features $\textbf{F}^G$ with different convolution layers. Then, the $Fuse$ operation combines them to obtain a compact feature representation $\textbf{F}^C$ by applying element-wise summation, global average pooling, and pixel-wise convolution, which can be formulated as:
\begin{equation}
    \textbf{F}^C = {\rm Conv}({\rm Pool}(\textbf{F}^R + \textbf{R}^G)).
\end{equation}
Finally, the $Select$ operation guides two other convolution layers followed by the softmax attention to enhance certain features: 
\begin{align}
    Attn^{R} &= {\rm SoftMax}({\rm Conv}^R(\textbf{F}^C), {\rm Conv}^G(\textbf{F}^C)), \\
    Attn^{G} &= {\rm SoftMax}({\rm Conv}^R(\textbf{F}^C), {\rm Conv}^G(\textbf{F}^C)),
\end{align}
where $Attn^{R}$ and $Attn^{G}$ represent the softmax attention of reconstruction and glyph features. The fused reconstruction and glyph features are computed by multiplying the softmax attention with the $\textbf{F}^R$ and the $\textbf{F}^G$, respectively:
\begin{align}
    \textbf{F}^{FR} &= Attn^{R} \textbf{F}^R, \\
    \textbf{F}^{FG} &= Attn^{G} \textbf{F}^G,
\end{align}
where $\textbf{F}^{FR}$ and $\textbf{F}^{FG}$ are fused reconstruction and glyph features. The fused features $\textbf{F}^F$ are obtained by summing the $\textbf{F}^{FR}$ and the $\textbf{F}^{FG}$:
\begin{equation}
    \textbf{F}^F = \textbf{F}^{FR} + \textbf{F}^{FG}.
\end{equation}

\subsection{Loss Function}

We train the model with PSNR loss for the reconstructed OBI image $\textbf{I}^{\prime}$ and perceptual loss for the reconstructed skeleton image $\textbf{S}^{\prime}$:
\begin{equation}
    \label{trade-off}
    \mathcal{L} = \alpha_1\mathcal{L}_1(\textbf{I}^{\prime}) + \alpha_2\mathcal{L}_2(\textbf{I}^{\prime}) + \alpha_3\mathcal{L}_1(\textbf{S}^{\prime}) + \alpha_4\mathcal{L}_2(\textbf{S}^{\prime}),
\end{equation}
where $\{\alpha_i\}_{i=1}^4$ are hyperparameters. $\mathcal{L}_1$ and $\mathcal{L}_2$ refer to PSNR and perceptual loss. Given an input image $\textbf{X}$, $\mathcal{L}_1$ and $\mathcal{L}_2$ are defined as:
\begin{align}
    \mathcal{L}_1(\textbf{X}) &= 10\log\left( \frac{\max^2(\textbf{X})}{{\rm MSE}(\textbf{X}_{GT}, \textbf{X})} \right), \\
    \mathcal{L}_2(\textbf{X}) &= \parallel {\rm VGG}(\textbf{X}_{GT}) - {\rm VGG}(\textbf{X}) \parallel_1,
\end{align}
where $\max$ and ${\rm MSE}$ denote the maximum value representing the color of the image pixel and the mean square error. $\textbf{X}_{GT}$ refers to the ground truth of the input image. $\mathcal{L}_2$ is computed based on a VGG16 \cite{simonyan2014very} model pre-trained on the ImageNet \cite{deng2009imagenet} dataset. The ground truth of the skeleton is obtained by an existing method \cite{jian2005chinese} based on mathematical morphology.

\section{Experiments}

In this section, we conduct comprehensive experiments on three representative OBI datasets, i.e., Oracle-50K \cite{han2020self}, RCRN \cite{shi2022rcrn}, and OBC306 \cite{huang2019obc306}, to evaluate the effectiveness of the proposed OBIFormer for the OBI denoising task. The details are as follows:

\subsection{Datasets}

\noindent\textbf{Oracle-50K:} The Oracle-50K dataset \cite{han2020self} is a large OBI dataset designed for OBI recognition and classification tasks. It contains various instances sourced from three different collections. The instances from Xiaoxuetang\footnote{http://xiaoxue.iis.sinica.edu.tw/jiaguwen} and Chinese Etymology\footnote{https://hanziyuan.net/} are gathered by a custom web-crawling tool, while other instances are generated with a TrueType font file. Considering the long-tail distribution of oracle character instances in the Oracle-50K dataset, we solely select the top 100 characters with the highest frequency for our experiments.

\noindent\textbf{RCRN:} The RCRN dataset \cite{shi2022rcrn} is sampled from historical Chinese character and oracle document datasets. Due to its complex real-world degradation, it is an essential benchmark for OBI denoising tasks \cite{shi2022rcrn, shi2022charformer}. While the test set is not publicly available, we use the training set for training, validation, and testing, which consists of 900 noisy-clean image pairs. For each pair, we split it into a noisy image and a clean image for experiments.

\noindent\textbf{OBC306:} The OBC306 dataset \cite{huang2019obc306} is a well-known rubbing dataset constructed using eight authoritative oracle bone publications worldwide. These publications are first scanned and encoded using a six-character/number code. The characters in scanned images are retrieved with the help of an oracle bone dictionary tool by comparing the text in the images with A List of Oracle Characters \cite{shen2008jiaguwen}. Finally, the best-matching character samples are added to the dataset.

\subsection{Implementation Details}

We use Oracle-50K, RCRN, and OBC306 datasets for our experiments. Among them, the RCRN and OBC306 datasets are rubbing datasets, and the Oracle-50K dataset is a handprint dataset. Specifically, we utilize STSN \cite{wang2022unsupervised} to apply the domain adaptation for the Oracle-50K dataset. Then, these datasets are used to train denoising models. Our model is implemented using the PyTorch platform and trained on an NVIDIA RTX 4090 GPU for 300 epochs with the AdamW optimizer \cite{loshchilov2017decoupled}. The learning rate is set to 2e-4, and the weight decay is set to 0.01. We use a batch size of 10, and all images are resized to 256 $\times$ 256. For the RCRN dataset, due to its limited data scale, we adopt a data augmentation technique consisting of random rotation and horizontal and vertical flipping. Considering various quality assessment metrics \cite{zhang2024quality, zhang2024bench, zhang2024q,chen2024study}, we apply two common metrics of low-level vision tasks, i.e., peak signal-to-noise ratio (PSNR) and the structural similarity index measure (SSIM) \cite{hore2010image}. Higher PSNR and SSIM values indicate better denoising results. For Oracle-50K and RCRN datasets, their target images serve as the ground truth to compute the PSNR and SSIM metrics. 

\subsection{Baseline Algorithms}

To evaluate the performance of the proposed OBIFormer, we select nine representative models, i.e., Denoising Convolutional Neural Networks (DnCNN) \cite{zhang2017beyond}, Dual Adversarial Network (DANet) \cite{yue2020dual}, Selective Residual M-Net (SRMNet) \cite{fan2022selective}, Kernel Basis Network (KBNet) \cite{zhang2023kbnet}, Invertible Denoising Network (InvDN) \cite{liu2021invertible}, CharFormer \cite{shi2022charformer}, U-Shaped Transformer (Uformer) \cite{wang2022uformer}, Restoration Transformer (Restormer) \cite{zamir2022restormer}, and CascadedGaze Network (CGNet) \cite{ghasemabadi2024cascadedgaze} for comparisons.

\subsection{Comparison with the State-of-the-Art}

\begin{table}[ht]
    \centering
    \caption{Quantitative comparisons of baseline methods and our OBIFormer on Oracle-50K \cite{han2020self} and RCRN \cite{shi2022rcrn} datasets. The compared methods include CNN, INN, GAN, and transformer-based models. The best and second-best results are in bold and underlined, respectively.}
    \label{results}
    \begin{tabular*}{\linewidth}{@{}CCCCC@{}}
        \toprule
        \toprule
        \multirow{2}{*}{Methods} & \multicolumn{2}{c}{Oracle-50K \cite{han2020self}} & \multicolumn{2}{c}{RCRN \cite{shi2022rcrn}} \\ 
        \cmidrule(r){2-3} \cmidrule(r){4-5} & PSNR$\uparrow$ & SSIM$\uparrow$ & PSNR$\uparrow$ & SSIM$\uparrow$ \\
        \midrule
        Raw Image & 10.25 & 0.099 & 9.111 & 0.611 \\
        DnCNN \cite{zhang2017beyond} & 14.31 & 0.690 & 20.40 & 0.794 \\
        DANet \cite{yue2020dual} & 12.11 & \underline{0.830} & 20.81 & \underline{0.950} \\
        SRMNet \cite{fan2022selective} & 14.17 & 0.820 & 21.96 & 0.945 \\
        KBNet \cite{zhang2023kbnet} & \underline{15.25} & 0.798 & \underline{22.03} & 0.942 \\
        InvDN \cite{liu2021invertible} & 12.30 & 0.815 & 18.96 & 0.941 \\
        CharFormer \cite{shi2022charformer} & 13.55 & \underline{0.830} & 19.57 & 0.924 \\
        Uformer \cite{wang2022uformer} & 13.64 & 0.797 & 21.44 & 0.893 \\
        Restormer \cite{zamir2022restormer} & \underline{15.25} & 0.797 & 21.59 & 0.945 \\
        CGNet \cite{ghasemabadi2024cascadedgaze} & 14.82 & 0.819 & 21.51 & 0.944 \\
                Ours & \textbf{16.31} & \textbf{0.893} & \textbf{22.19} & \textbf{0.969} \\
        \bottomrule
        \bottomrule
    \end{tabular*}
\end{table}

\begin{figure*}[t]
    \centering
    \subfigure{
        \includegraphics[width=\linewidth]{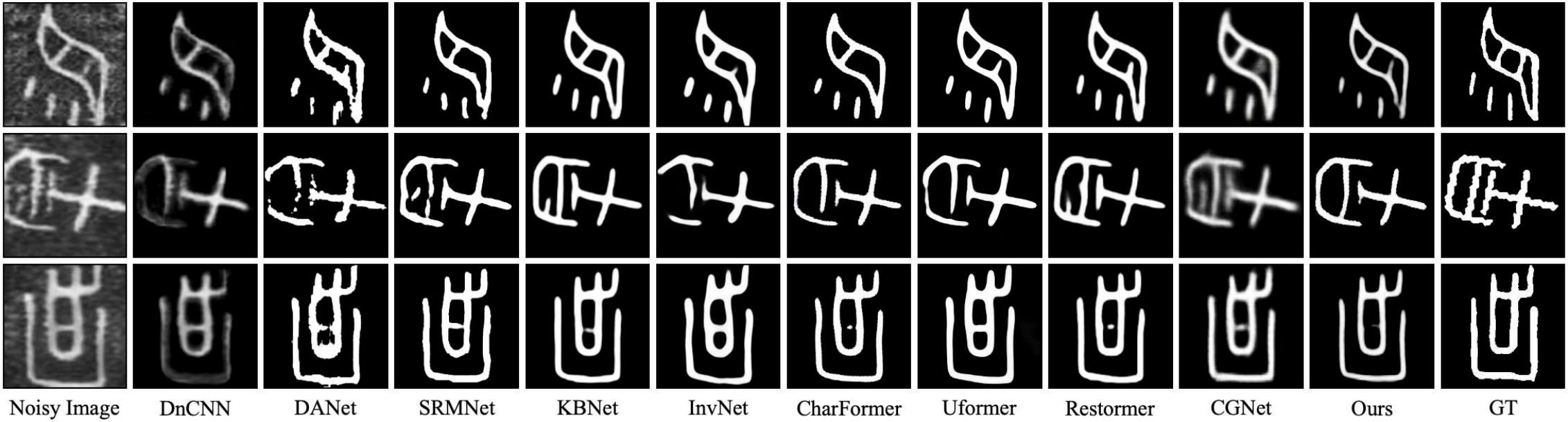}
    }
    \caption{Qualitative comparisons of baseline methods and our OBIFormer on Oracle-50K dataset \cite{han2020self}.}
    \label{vis_oracle-50k}
\end{figure*}

\begin{figure*}[t]
    \centering
    \subfigure{
        \includegraphics[width=\linewidth]{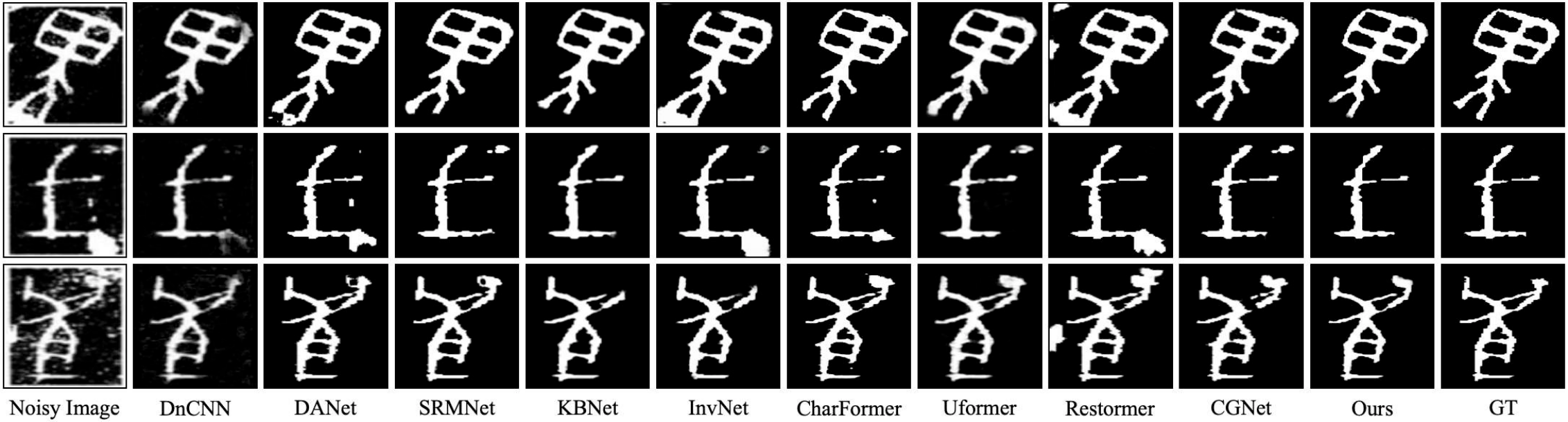}
    }
    \caption{Qualitative comparisons of baseline methods and our OBIFormer on RCRN dataset \cite{shi2022rcrn}.}
    \label{vis_dataset2}
\end{figure*}

Table \ref{results} reports the quantitative comparisons of baseline methods and our OBIFormer on Oracle-50K and RCRN datasets. On the Oracle-50K dataset, OBIFormer achieves 16.31 dB on PSNR and 0.893 on SSIM, surpassing all the other methods by at least 1.06 dB and 0.063, respectively. Similarly, on the RCRN dataset, OBIFormer attains 22.19 dB on PSNR and 0.969 on SSIM, outperforming all the other methods by at least 0.16 dB in PSNR and 0.019 in SSIM. Besides, DANet achieves 0.830 and 0.950 in terms of SSIM on Oracle-50K and RCRN, respectively. KBNet obtains PSNR values of 15.25 on Oracle-50K and 22.03 on RCRN. However, these methods excel in only one metric while exhibiting poor performance in the other. In contrast, our OBIFormer achieves state-of-the-art performance on both PSNR and SSIM, with a particularly notable improvement on SSIM, highlighting the effectiveness of GSNB and SKFF.

Furthermore, we provide qualitative comparisons of baseline methods and our OBIFormer on Oracle-50K and RCRN datasets. As illustrated in Fig. \ref{vis_oracle-50k} and Fig. \ref{vis_dataset2}, OBIFormer effectively removes complex noise while preserving glyph details, whereas other methods struggle to maintain glyph consistency. Notably, OBIFormer generates cleaner and visually closer images to the ground truths than other algorithms in challenging scenarios where noise and strokes are difficult to distinguish. For example, most other methods fail to remove the spindle-shaped noise in the second case in Fig. \ref{vis_dataset2}. In the first case in Fig. \ref{vis_oracle-50k} and the third case in Fig. \ref{vis_dataset2}, only our OBIFormer succeeds in restoring the broken strokes precisely. Overall, OBIFormer achieves state-of-the-art performance on Oracle-50K and RCRN quantitatively and qualitatively.

\subsection{OBI Recognition}

To further validate the effectiveness of OBI denoising in enhancing recognition accuracy, we employ ResNet-18, ResNet-50, and ResNet-152 \cite{he2016deep} for the OBI recognition task on the test set of the Oracle-50K dataset. The test set is divided into training and testing subsets in a 7:3 ratio. The model is pre-trained on the ImageNet dataset \cite{deng2009imagenet} and fine-tuned on the Oracle-50K dataset for 100 epochs using the Adam optimizer. We adopt a data augmentation technique of random rotation. The learning rate is set to 1e-3 (5e-4 for ResNet-152) with a batch size of 256 (128 for ResNet-50, 64 for ResNet-152). We compare the original Oracle-50K dataset with the denoising results generated by OBIFormer trained on the same dataset.

\begin{figure}[t]
    \centering
    \includegraphics[width=\linewidth]{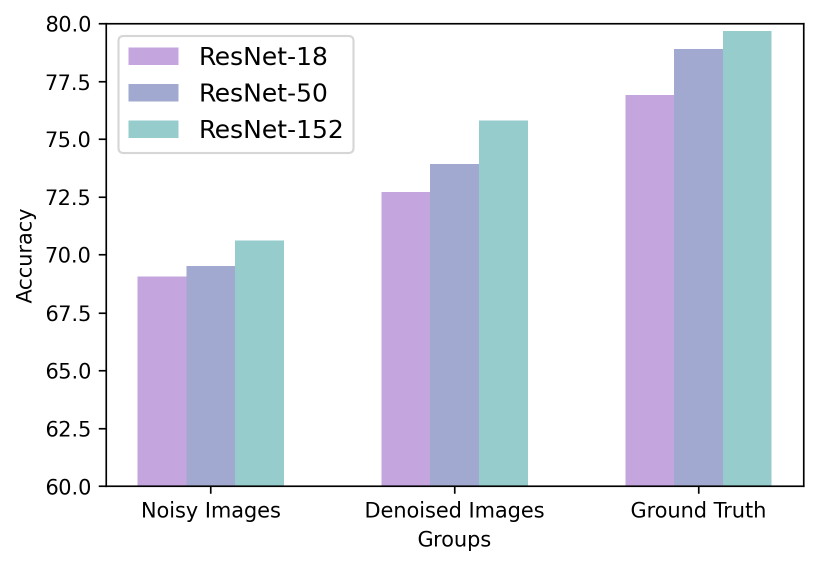}
    \caption{Recognition results of ResNet-18, ResNet-50, and ResNet-152 \cite{he2016deep} on Oracle-50K dataset \cite{han2020self}.}
    \label{rec}
\end{figure}

As illustrated in Fig. \ref{rec}, ResNet-18 obtains a significant gain of 3.65\% in the recognition accuracy on denoised images compared to the noisy images, demonstrating the effectiveness of OBI denoising in improving recognition accuracy. In addition, we observe a more remarkable improvement (4.42\% and 5.19\%) when deeper networks were applied, i.e., ResNet-50 and ResNet-152. This is attributed to the fact that denoised images contain more discriminative features, which are better captured by deeper networks. However, the recognition accuracy of the denoised image is still far from the ground truth.

\subsection{Computational efficiency}

\begin{table*}[t]
    \centering
    \fontsize{8pt}{10pt}\selectfont
    \caption{The number of parameters (\#Param.), FLOPs, and inference time of baseline methods and our OBIFormer.}
    \label{flops}
    \begin{tabular}{ccccccccccc}
	\toprule
	\toprule
	& DnCNN & DANet & SRMNet & KBNet & InvDN & CharFormer & Uformer & Restormer & CGNet & Ours \\
        \midrule
        \#Param. (M) & 0.67 & 63.01 & 37.59 & 104.93 & 2.64 & 13.10 & 50.88 & 26.10 & 119.22 & 8.35 \\
        FLOPs (G) & 44.02 & 32.73 & 285.17 & 14.47 & 95.08 & 51.04 & 89.46 & 140.99 & 62.10 & 20.45  \\
        Inference Time (ms) & 1.94 & 1.19 & 15.73 & 36.60 & 3.04 & 19.92 & 19.23 & 46.34 & 10.88 & 10.35 \\
	\bottomrule
	\bottomrule
    \end{tabular}
\end{table*}

Table \ref{flops} presents comprehensive comparisons of the computational efficiency of baseline methods and our OBIFormer. We evaluate the number of parameters (\#Param.), FLOPs, and inference time. To ensure a consistent comparison, \#Param., FLOPs, and inference time are calculated based on a random input patch with a 256 $\times$ 256 resolution. We perform 50 iterations of GPU warm-up for the inference time before the measurement. Consequently, OBIFormer has fewer FLOPs than most baseline methods. Compared to CharFormer, OBIFormer has 3.02$\times$ fewer parameters and runs 4.76$\times$ faster. When deployed on the 13th Gen Intel(R) Core(TM) i9-13900K CPU @ 3.00GHz, 32GB RAM, and an NVIDIA GeForce RTX 4090 GPU for acceleration, OBIFormer processes an image in just 10.35 ms, yielding competitive inference efficiency.

\subsection{Exploration of the Generalization Ability}

\begin{figure}[t]
    \centering
    \includegraphics[width=\linewidth]{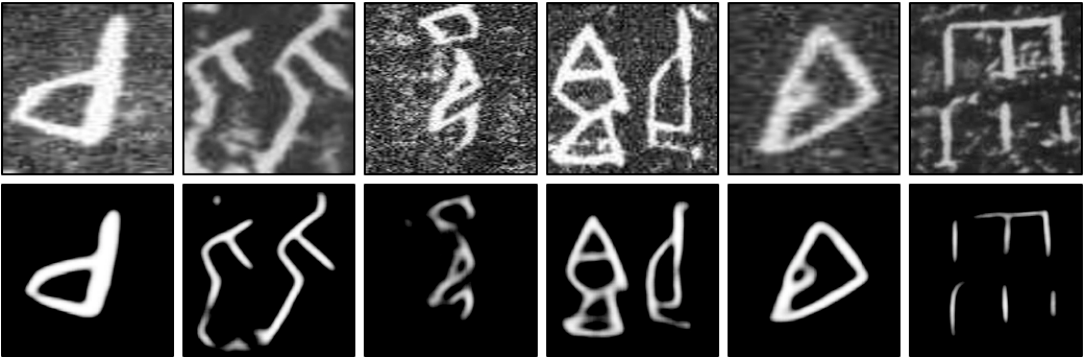}
    \caption{Denoising results of our OBIFormer (trained on Oracle-50K dataset \cite{han2020self}) on OBC306 dataset \cite{huang2019obc306}.}
    \label{vis_Oracle-50K_OBC306}
\end{figure}

\begin{figure}[t]
    \centering
    \includegraphics[width=\linewidth]{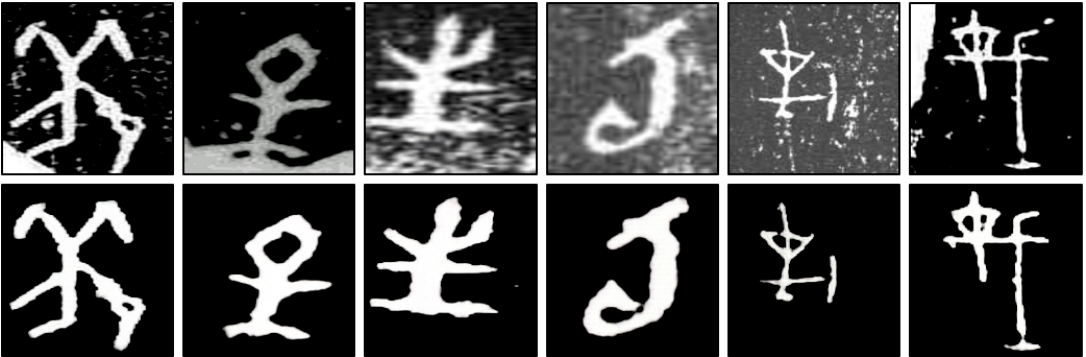}
    \caption{Denoising results of our OBIFormer (trained on RCRN dataset \cite{shi2022rcrn}) on OBC306 dataset \cite{huang2019obc306}.}
    \label{vis_dataset2_OBC306}
\end{figure}

To explore the generalization ability of our OBIFormer, we test it on a real oracle dataset (i.e., the OBC306 dataset) after training it on Oracle-50K and RCRN datasets. As shown in Fig. \ref{vis_Oracle-50K_OBC306} and Fig. \ref{vis_dataset2_OBC306}, OBIFormer shows a strong generalization ability on the OBC306 dataset. On the one hand, the denoising results of OBIFormer trained on the Oracle-50K dataset demonstrate the effectiveness of the adapted texture, which can be easily obtained with STSN or other OBI generation methods. On the other hand, even when we train OBIFormer on the RCRN dataset, which contains only 900 noisy-clean image pairs, it still performs well on the OBC306 dataset. Therefore, with a large amount of synthetic noisy-clean image pairs of OBIs, the generalization ability of OBIFormer can be further improved. This study demonstrates the great potential of OBIFormer in assisting automatic OBI recognition.

\subsection{Ablation Studies}

To validate the effectiveness of our OBIFormer, we conduct ablation studies on the RCRN dataset. In our experiments, we analyze the contribution of each core component in OBIFormer and the impact of specific hyperparameters.

\begin{figure*}[t]
    \centering
    \includegraphics[width=\linewidth]{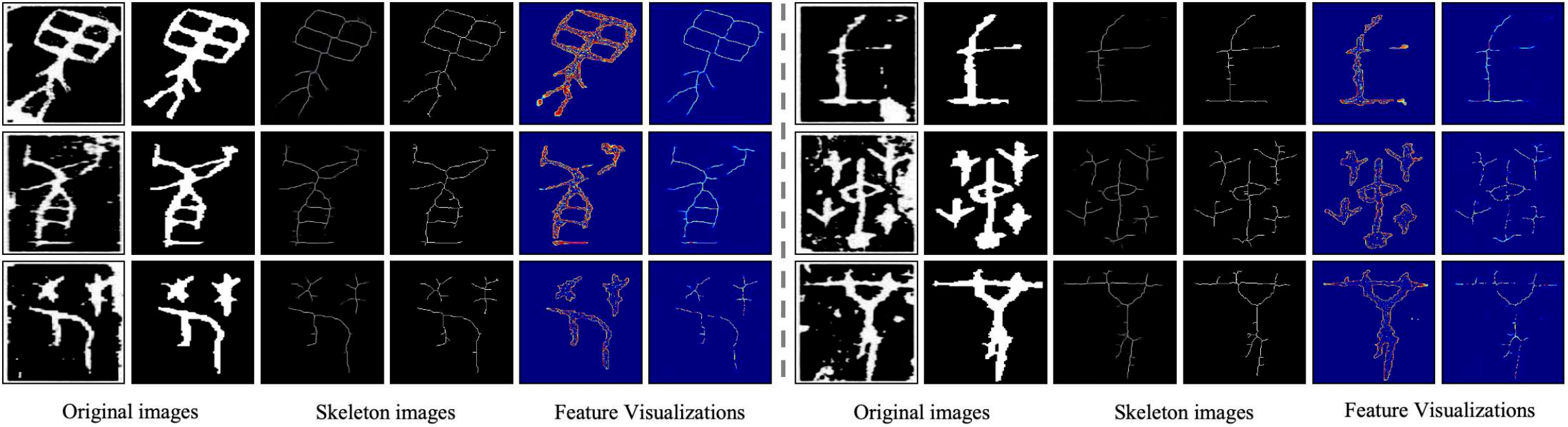}
    \caption{Visualization results of our OBIFormer on RCRN dataset \cite{shi2022rcrn}. For each case, the first two images are the noisy character image and its ground truth, the second two refer to the reconstructed skeleton image and its ground truth, and the last two are the visualization of reconstruction and glyph features.}
    \label{vis}
\end{figure*}

\noindent\textbf{Validation of Glyph Information Extraction.} To demonstrate that glyph structural network blocks (GSNBs) can effectively extract glyph information from input images, we visualize the output of the additional feature corrector. The results can be found in Fig. \ref{vis}. For each case in Fig. \ref{vis}, the first two images refer to the noisy character image and its ground truth, and the second two are the reconstructed skeleton image and its ground truth, where we can find that the glyphs are extracted properly. This visualization indicates that our OBIFormer can precisely extract glyph features from the input images.

\begin{table}[t]
    \centering
    \caption{Effects of different feature fusion strategies.}
    \label{skff}
    \begin{tabular}{cccc}
	\toprule
	\toprule
	Strategies & Addition & Concatenation & SKFF \\
        \midrule
        PSNR$\uparrow$ & 20.69 & 20.96 & 22.19 \\
        SSIM$\uparrow$ & 0.962 & 0.967 & 0.969 \\
	\bottomrule
	\bottomrule
    \end{tabular}
\end{table}

\noindent\textbf{Evaluation of Feature Fusion Strategy.} To evaluate the effectiveness of the selective kernel feature fusion (SKFF) module, we apply simple addition and concatenation of the output of $i$-th CASB and GSNB for comparison. As illustrated in Table \ref{skff}, the SKFF module provides favorable gains of 1.50 dB and 1.23 dB compared to simple addition and concatenation in terms of PSNR. Similar performance gains can be observed in the SSIM metric. That's because the SKFF module can generate attention maps for the reconstruction and glyph features and aggregate them dynamically. Since the glyph features can guide the model in reconstructing the denoised image, a more significant improvement in PSNR is obtained compared to SSIM.

\begin{figure}[t]
    \centering
    \subfigure[$\alpha_1$]{
        \includegraphics[width=0.48\linewidth]{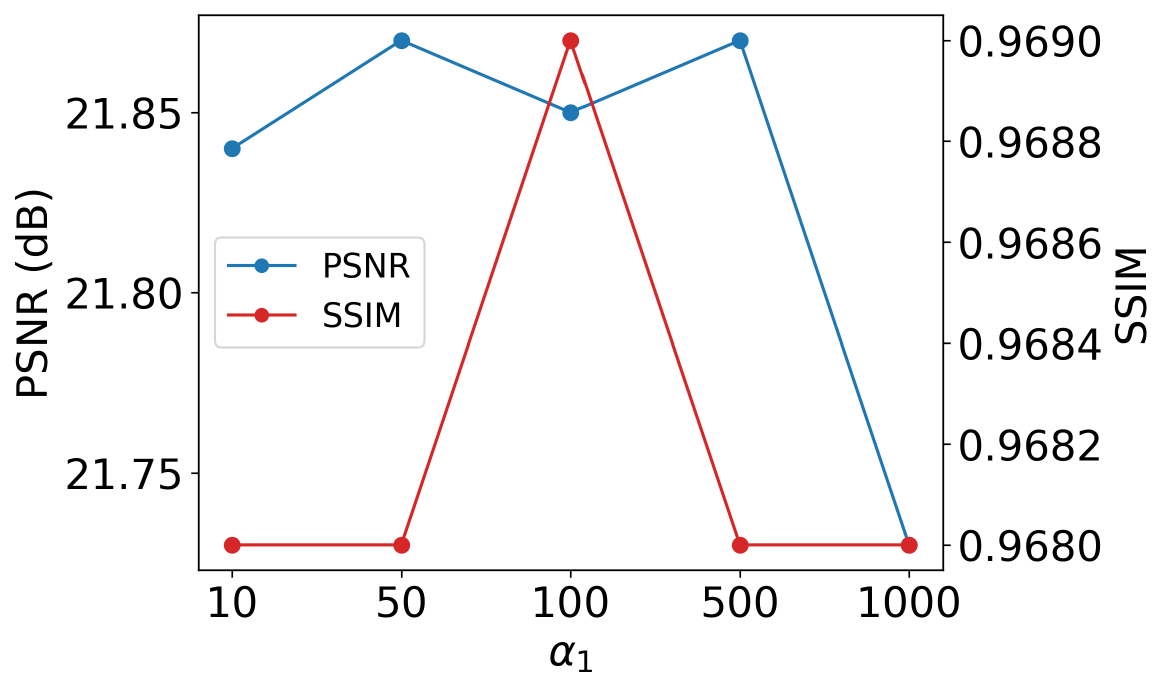}
        \hspace{-0.05\linewidth} 
    }
    \subfigure[$\alpha_2$]{
        \includegraphics[width=0.48\linewidth]{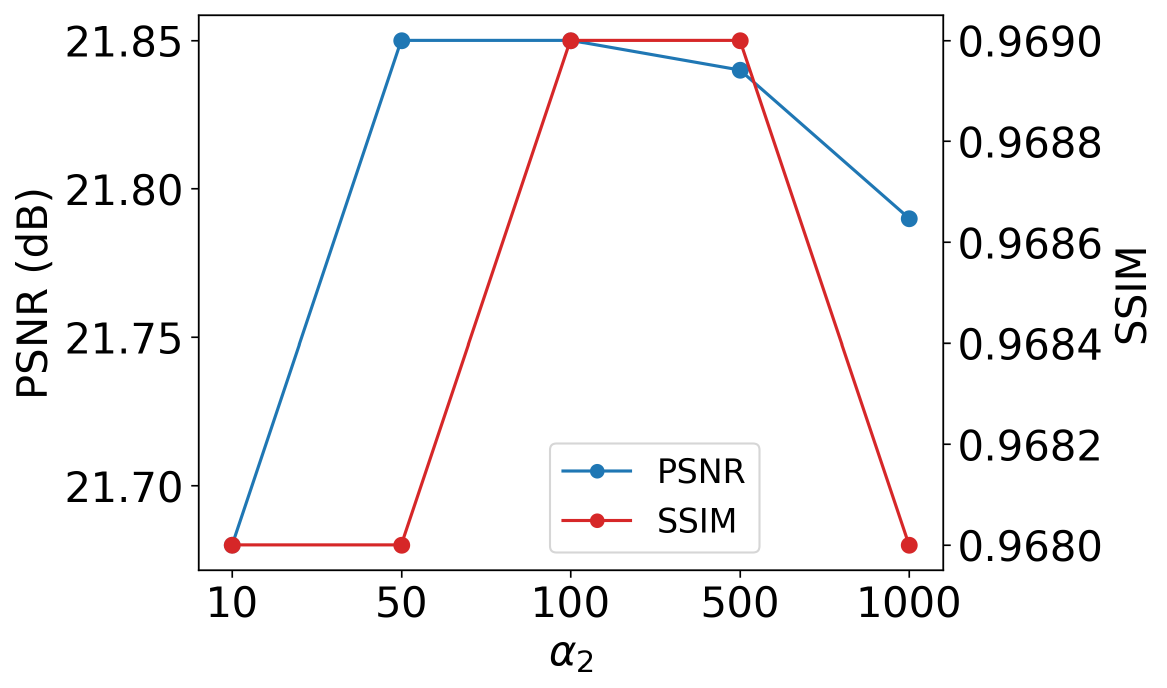}
    }
    \subfigure[$\alpha_3$]{
        \includegraphics[width=0.48\linewidth]{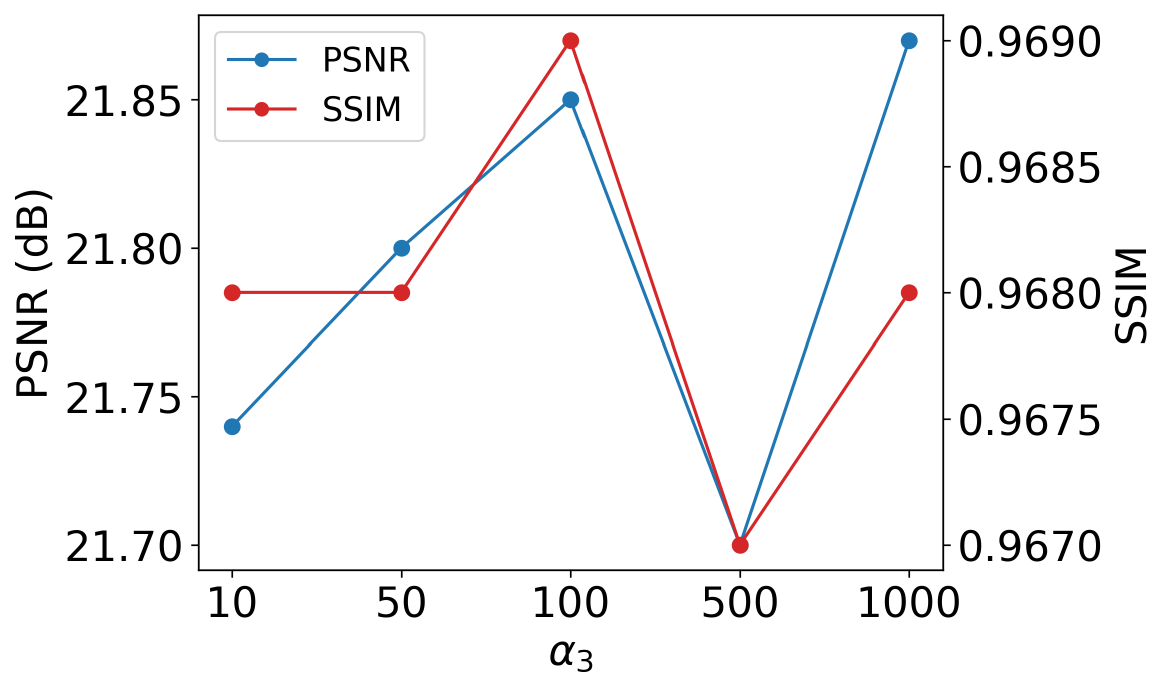}
        \hspace{-0.05\linewidth} 
    }
    \subfigure[$\alpha_4$]{
        \includegraphics[width=0.48\linewidth]{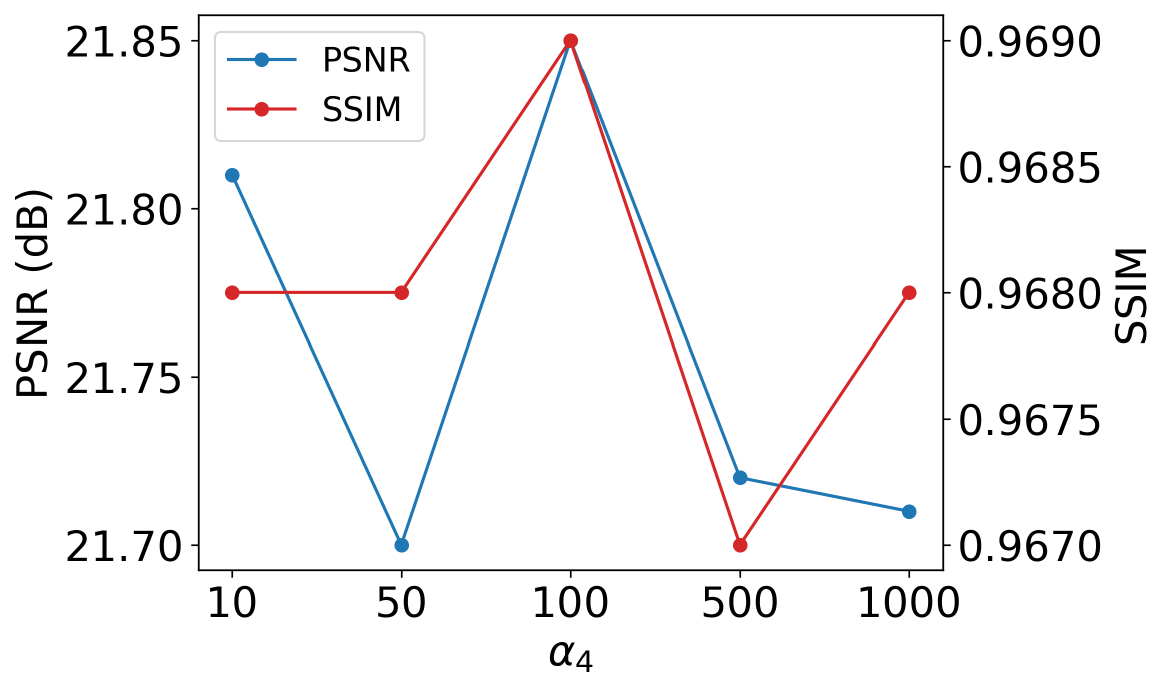}
    }
    \caption{The sensitivity of PSNR and SSIM to $\{\alpha_i\}_{i=1}^4$.}
    \label{parameters}
\end{figure}

\noindent\textbf{Impact of the Hyperparameters.} $\{\alpha_i\}_{i=1}^4$ are the hyperparameters for different losses in Eq. \ref{trade-off}. To investigate the performance of OBIFormer when these parameters change, we conduct ablation experiments and show the results in Fig. \ref{parameters}. The models are trained for 200 epochs in all experiments for computational reasons. It can be observed that PSNR and SSIM metrics first increase and then decrease as $\alpha_1$ and $\alpha_2$ vary, demonstrating a desirable bell-shaped curve. In principle, small values of $\alpha_1$ and $\alpha_2$ (e.g., 10) would limit the performance of the CSABs, while large values of $\alpha_1$ and $\alpha_2$ (e.g., 1000) would weaken the effect of GSNBs. We also observe that $\alpha_3$ and $\alpha_4$ lead to similar PSNR and SSIM metrics trends, indicating that glyph features can guide the model in reconstructing the denoised image.

\begin{figure}[t]
    \centering
    \subfigure[PSNR Metric]{
        \includegraphics[width=0.48\linewidth]{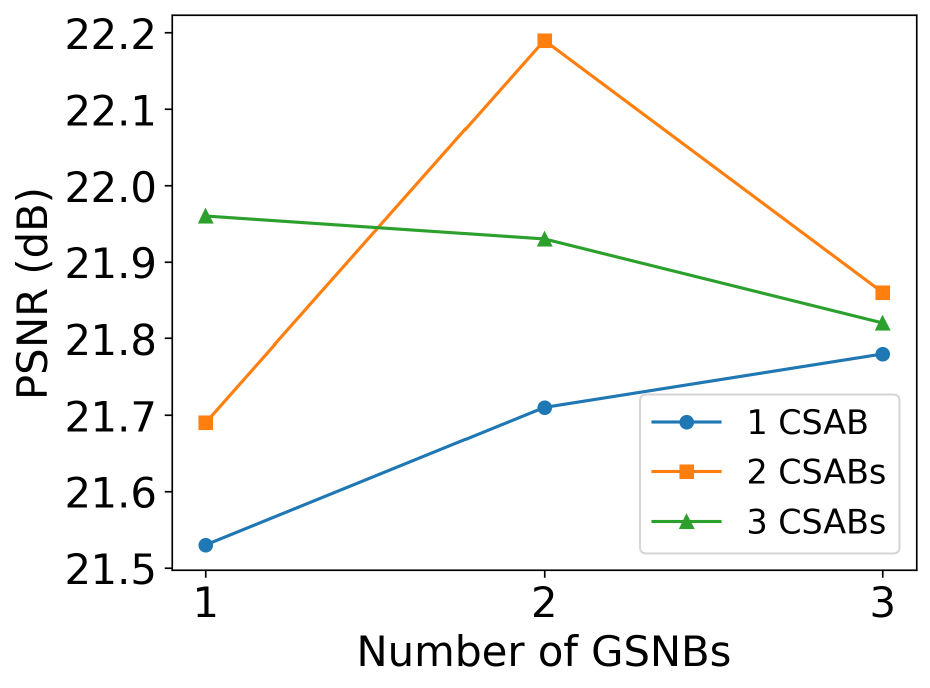}
        \hspace{-0.05\linewidth} 
    }
    \subfigure[SSIM Metric]{
        \includegraphics[width=0.48\linewidth]{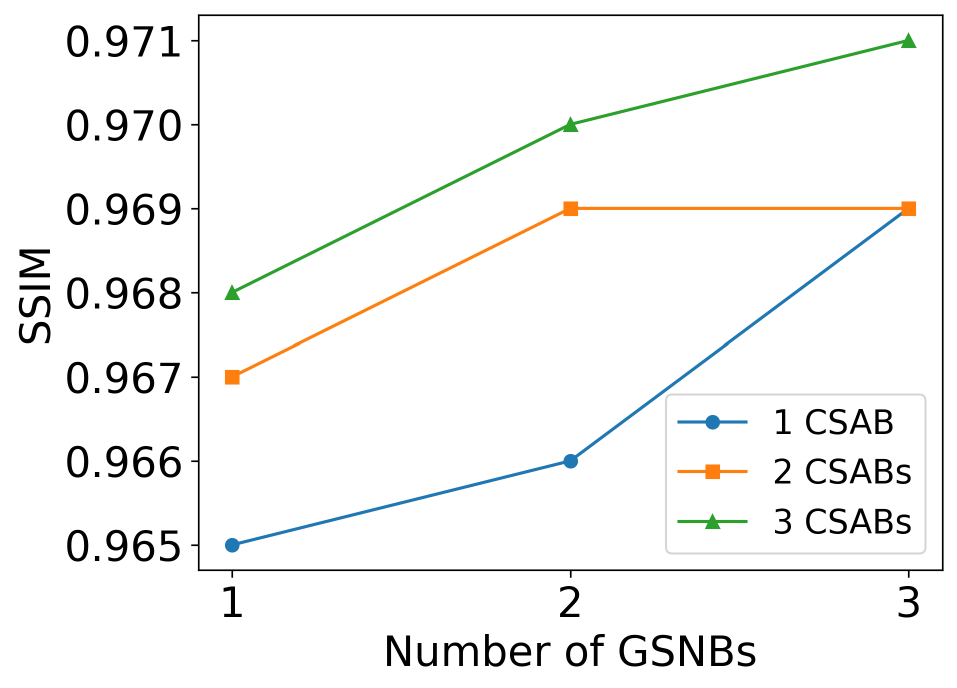}
    }
    \caption{Effects of different CSABs and GSNBs settings.}
    \label{CSAB&GSNB}
\end{figure}

Additionally, we adjust the number of CSABs and GSNBs to optimize the performance of our OBIFormer. Fig. \ref{CSAB&GSNB} shows the effects of different CSABs and GSNBs settings. As the number of CSABs and GSNBs increases, the PSNR metric initially rises but eventually declines due to overfitting as model complexity grows. In contrast, the SSIM metric continues to improve even when the model is overfitted. That can be attributed to the SKFF module, which effectively aggregates the overfitted reconstruction features and glyph features, mitigating the negative influence of excessive CSABs. Consequently, we use two CSABs and GSNBs in this paper.

\subsection{Visualization}

To further demonstrate the effectiveness of our OBIFormer, we conduct visualization studies. As shown in Fig. \ref{vis}, we visualize the deep features extracted by the final OFB, which consists of reconstruction and glyph features. For each case in Fig. \ref{vis}, the last two images correspond to the visualizations of reconstruction and glyph features, respectively. For the reconstruction features, OBIFormer successfully captures character-related features while removing complex noise. For the glyph features, OBIFormer precisely learns the glyph information from the input images, aided by the GSNBs and SKFFs. These visualizations indicate that both reconstruction and glyph features are successfully learned between the OFBs.

\section{Conclusion and Future Work}

In this paper, we propose a fast attentive denoising framework for OBIs, i.e., OBIFormer. Specifically, our OBIFormer consists of an input projector, an output projector, an additional feature corrector, and several OFBs. The OFB utilizes CSABs to extract reconstruction features and GSNBs to learn glyph information. Additionally, the SKFF module aggregates reconstruction features and glyph information dynamically. Extensive experiments demonstrate the superiority of OBIFormer on Oracle-50K and RCRN datasets quantitatively and qualitatively. Furthermore, OBIFormer shows a strong generalization ability on the OBC306 dataset, demonstrating its great potential in assisting automatic OBI recognition. Finally, we provide comprehensive ablations to validate the effectiveness of each module.

However, the performance of OBIFormer on OBIs with some types of noise, such as bone-cracked and dense white regions, remains unsatisfactory. That is because there are only a few images with such noise, which can be treated as a few-shot learning problem. Hence, one future trend is to investigate more generic methods, e.g., the conditional diffusion model, to utilize multi-modal conditions to generate a noise-balanced dataset for OBI denoising. Also, we can apply a conditional diffusion model for image denoising directly, where conditions guide the model in removing target noise.

\section*{Acknowledgement}

This work was supported by the National Social Science Foundation of China (24Z300404220) and the Shanghai Philosophy and Social Science Planning Project (2023BYY003).

\balance
\clearpage

\bibliographystyle{unsrt}
\bibliography{displays}

\end{document}